\documentclass{article}

\usepackage[preprint]{neurips_2026}
\usepackage[utf8]{inputenc} 
\usepackage[T1]{fontenc}    
\usepackage{hyperref}       
\usepackage{url}            
\usepackage{booktabs}       
\usepackage{amsfonts}       
\usepackage{nicefrac}       
\usepackage{microtype}      
\usepackage[table]{xcolor}  
\usepackage{graphicx}
\usepackage{multirow}
\usepackage{makecell}
\usepackage{float}
\usepackage{amsmath}
\usepackage{amssymb}
\usepackage{tabularx}
\usepackage{array}

\title{\texorpdfstring{CoT$^2$-Meta: Budgeted Metacognitive Control for Test-Time Reasoning}{CoT2-Meta: Budgeted Metacognitive Control for Test-Time Reasoning}}

\author{%
  Siyuan Ma \\
  Nanyang Technological University \\
  \texttt{MASI0004@e.ntu.edu.sg}
  \And
  Bo Gao \\
  Carnegie Mellon University
  \And
  Zikai Xiao \\
  Zhejiang University
  \AND
  Hailong Wang \\
  Sun Yat-Sen University
  \And
  Xinlei Yu \\
  National University of Singapore
  \And
  Rui Qian \\
  Fudan University
  \AND
  Jiayu Qian \\
  City University of Hong Kong (Dongguan)
  \And
  Luqi Gong \\
  Zhejiang Lab
  \And
  Yang Liu \\
  Nanyang Technological University
}
\begin{document}

\maketitle

\begin{abstract}
Recent test-time reasoning methods improve performance by generating more candidate chains or searching over larger reasoning trees, but they typically lack explicit control over \emph{when} to expand, \emph{what} to prune, \emph{how} to repair, and \emph{when} to abstain.
We introduce \textsc{CoT$^2$-Meta}, a training-free metacognitive reasoning framework that combines object-level chain-of-thought generation with meta-level control over partial reasoning trajectories.
The framework integrates four components: strategy-conditioned thought generation, tree-structured search, an online process oracle for step-level reasoning evaluation, and a meta-controller that allocates computation through expansion, pruning, repair, stopping, and fallback decisions.

Under matched inference budgets, \textsc{CoT$^2$-Meta} consistently outperforms strong single-path, sampling-based, and search-based baselines, including ReST-MCTS.
On the default backbone, it achieves 92.8 EM on \textsc{MATH}, 90.4 accuracy on \textsc{GPQA}, 98.65 EM on \textsc{GSM8K}, 75.8 accuracy on \textsc{BBEH}, 85.6 accuracy on \textsc{MMMU-Pro}, and 48.8 accuracy on \textsc{HLE}, with gains over the strongest non-\textsc{CoT$^2$-Meta} baseline of +3.6, +5.2, +1.15, +2.0, +4.3, and +4.3 points, respectively.
Beyond these core results, the framework remains effective across a broader 15-benchmark suite spanning knowledge and QA, multi-hop reasoning, coding, and out-of-distribution evaluation.

Beyond aggregate accuracy, we show that the gains are not reducible to brute-force compute: \textsc{CoT$^2$-Meta} yields better compute scaling, improved calibration, stronger selective prediction, and targeted repair behavior under token-matched comparisons.
Additional analyses show that the framework generalizes across backbone families while decision-trace audits and failure taxonomies reveal interpretable controller behavior and localized remaining failure modes.
These results suggest that explicit metacognitive control is a practical design principle for reliable and compute-efficient test-time reasoning systems.
\end{abstract}
\section{Introduction}

Large language models (LLMs) benefit substantially from additional test-time compute on complex reasoning tasks. Chain-of-thought (CoT) prompting, self-consistency, and related scaling strategies show that more inference-time reasoning can improve final performance \cite{26,27}. More recent methods extend single-path reasoning to structured exploration, including tree-based search, graph-based search, and process-reward-guided deliberation \cite{14,2,1}. However, most still treat extra compute mainly as \emph{more generation} rather than \emph{better control}: they expand search more broadly, but typically do not explicitly decide when a partial trajectory should be expanded, pruned, repaired, or terminated \cite{14,1}. As a result, computation is often wasted on weak branches, while fluent but incorrect trajectories may persist too long.

A parallel line of work shows that reliable reasoning requires more than final-answer correctness. Verifier-based methods and process supervision highlight the importance of intermediate reasoning quality \cite{28,29,6,17}; related studies show that outcome accuracy can mask flawed reasoning processes \cite{6}, while LLM confidence is often poorly calibrated unless explicitly modeled or controlled \cite{21,22,20}. More broadly, metareasoning and computation-selection research frames deliberation as a decision problem under limited resources: the question is not only how to reason, but also which computations are worth performing, and when \cite{31,32}. These observations motivate systems that reason not only through trajectories, but also about how reasoning should proceed.

\begin{figure*}[!t]
  \centering
  \includegraphics[width=0.88\textwidth]{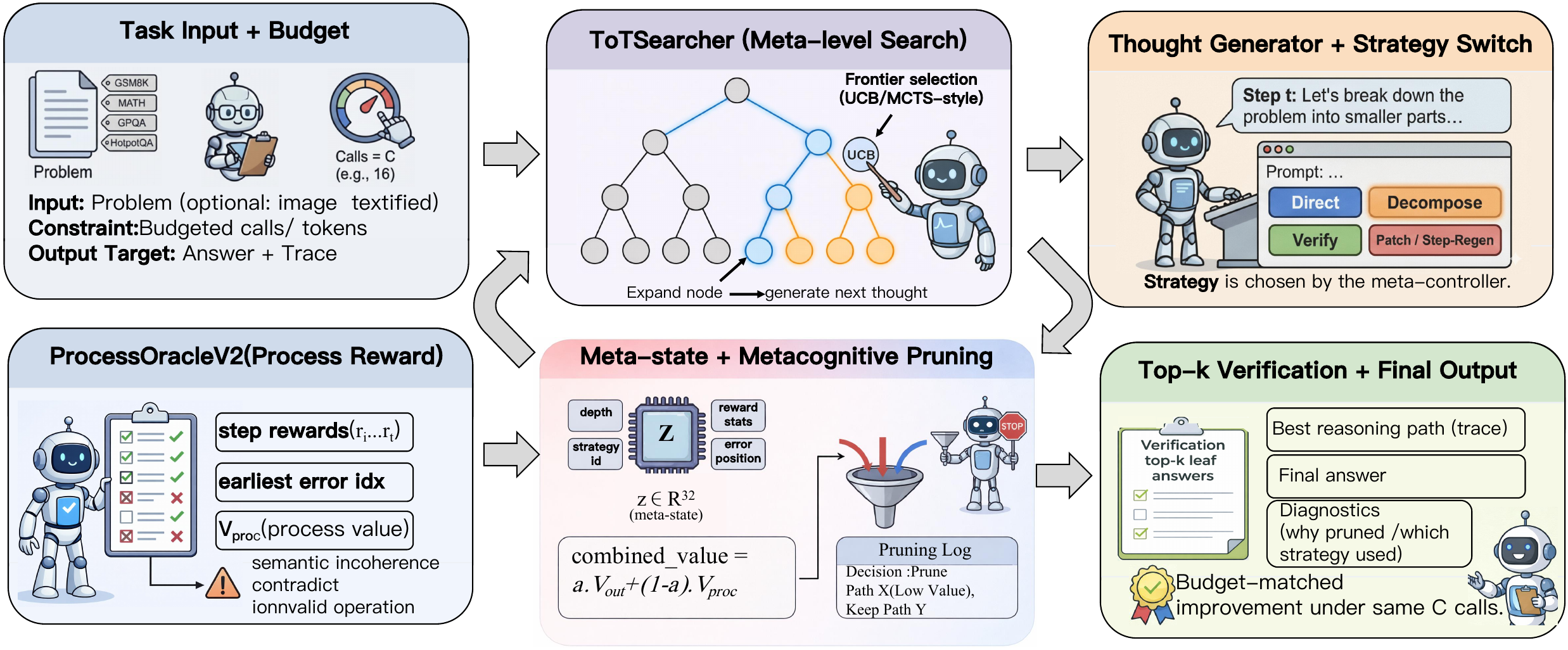}
  \caption{\textbf{CoT$^2$-Meta as reasoning over reasoning.} The inner layer performs object-level reasoning, while the outer layer performs metacognitive control over partial trajectories. The meta-controller decides when to expand, prune, repair, stop, or abstain under a bounded budget.}
  \label{fig:cot2meta_teaser}
\end{figure*}
We introduce \textsc{CoT$^2$-Meta}, a budgeted metacognitive reasoning framework that separates \emph{object-level reasoning} from \emph{meta-level control}. An LLM generates candidate reasoning steps that form a tree of partial trajectories, while a controller evaluates process quality, maintains a compact meta-state, and decides whether to \emph{expand}, \emph{prune}, \emph{repair}, \emph{stop}, or \emph{abstain}. This casts test-time reasoning as sequential control over trajectories rather than passive sample-and-rerank.

Our design follows two principles. First, useful test-time scaling should be \emph{selective}: computation should be allocated according to intermediate quality and uncertainty rather than spent uniformly across trajectories \cite{25,35,38}. Second, confidence should be \emph{actionable}: the same signals used to assess trustworthiness should also govern whether the system continues searching, invokes repair, or declines to answer. Accordingly, \textsc{CoT$^2$-Meta} integrates process evaluation, meta-state construction, confidence-aware routing, and decision logging into a single inference-time loop.

Empirically, \textsc{CoT$^2$-Meta} consistently outperforms strong baselines under matched compute budgets. On the default backbone, it improves over the strongest non-\textsc{CoT$^2$-Meta} baseline by $+3.6$ on \textsc{MATH}, $+5.2$ on \textsc{GPQA}, $+1.15$ on \textsc{GSM8K}, $+2.0$ on \textsc{BBEH}, $+4.3$ on \textsc{MMMU-Pro}, and $+4.3$ on \textsc{HLE}. The method also remains effective across a broader 15-benchmark suite spanning knowledge and QA, multi-hop QA, coding, and out-of-distribution evaluation, with gains that persist across backbone families under compute-normalized evaluation and are accompanied by better calibration, stronger selective prediction, more effective repair, and interpretable decision traces.

\noindent\textbf{Contributions.}
(1) We introduce \textsc{CoT$^2$-Meta}, a unified inference-time framework for process-aware metacognitive control over reasoning trajectories.
(2) We show consistent gains across a 15-benchmark suite spanning reasoning, multimodal reasoning, knowledge and QA, multi-hop QA, coding, and out-of-distribution evaluation under strict compute normalization.
(3) We show that these gains arise from stronger reasoning control rather than brute-force extra compute alone.
\section{Method}

We present \textsc{CoT$^2$-Meta}, a budgeted metacognitive reasoning framework that explicitly separates \emph{object-level reasoning} from \emph{meta-level control}.
At the object level, a backbone language model generates candidate reasoning steps.
At the meta level, a controller evaluates partial trajectories, maintains a compact state representation, and decides whether to expand, prune, repair, stop, or abstain.
This design turns test-time reasoning from pure text generation into a sequential control problem over reasoning trajectories.

\subsection{Problem Formulation}

Let $x$ denote an input problem.
A reasoning trajectory is a sequence of intermediate thoughts
\[
\tau_t = (z_1, z_2, \dots, z_t),
\]
where each $z_i$ is a coherent reasoning unit such as a sub-derivation, decomposition step, verification step, or intermediate conclusion.
The final output is either an answer $\hat{y}$ or an abstention decision $\varnothing$.

We assume access to an object-level generator
\[
G_\theta(z_{t+1} \mid x, \tau_t, s_t),
\]
where $s_t$ denotes the current control context, including the active reasoning strategy and any meta-level signals exposed to the generator.
Unlike standard chain-of-thought prompting, which follows a single trajectory, we consider a search space over multiple trajectories organized as a reasoning tree.

Let $\mathcal{T}$ denote the evolving reasoning tree and let $\mathcal{F}_t$ be its frontier at step $t$.
Each frontier node corresponds to a partial reasoning trajectory.
The system operates under a finite inference budget $C$, which counts all generation, evaluation, repair, and control calls.
The goal is to maximize answer quality while respecting this budget:
\[
\max_{\pi} \; \mathbb{E}\!\left[ U(\hat{y}, x) \right]
\quad \text{s.t.} \quad
\mathrm{Cost}(\pi; x) \le C,
\]
where $\pi$ is the meta-control policy and $U(\hat{y},x)$ is a utility that rewards correct high-confidence answers and can optionally penalize unsafe overconfident predictions.

This formulation differs from conventional test-time scaling in two ways.
First, the system does not spend compute uniformly across all trajectories.
Second, the controller can choose not only \emph{which branch to expand}, but also \emph{whether further reasoning is worthwhile at all}.
Hence, the problem is not merely ``generate more thoughts,'' but rather:
\emph{under a limited budget, how should computation be allocated across generation, evaluation, repair, and stopping decisions?}

\subsection{\texorpdfstring{CoT$^2$-Meta Framework}{CoT2-Meta Framework}}

\begin{figure*}[!t]
  \centering
  \includegraphics[width=0.80\textwidth]{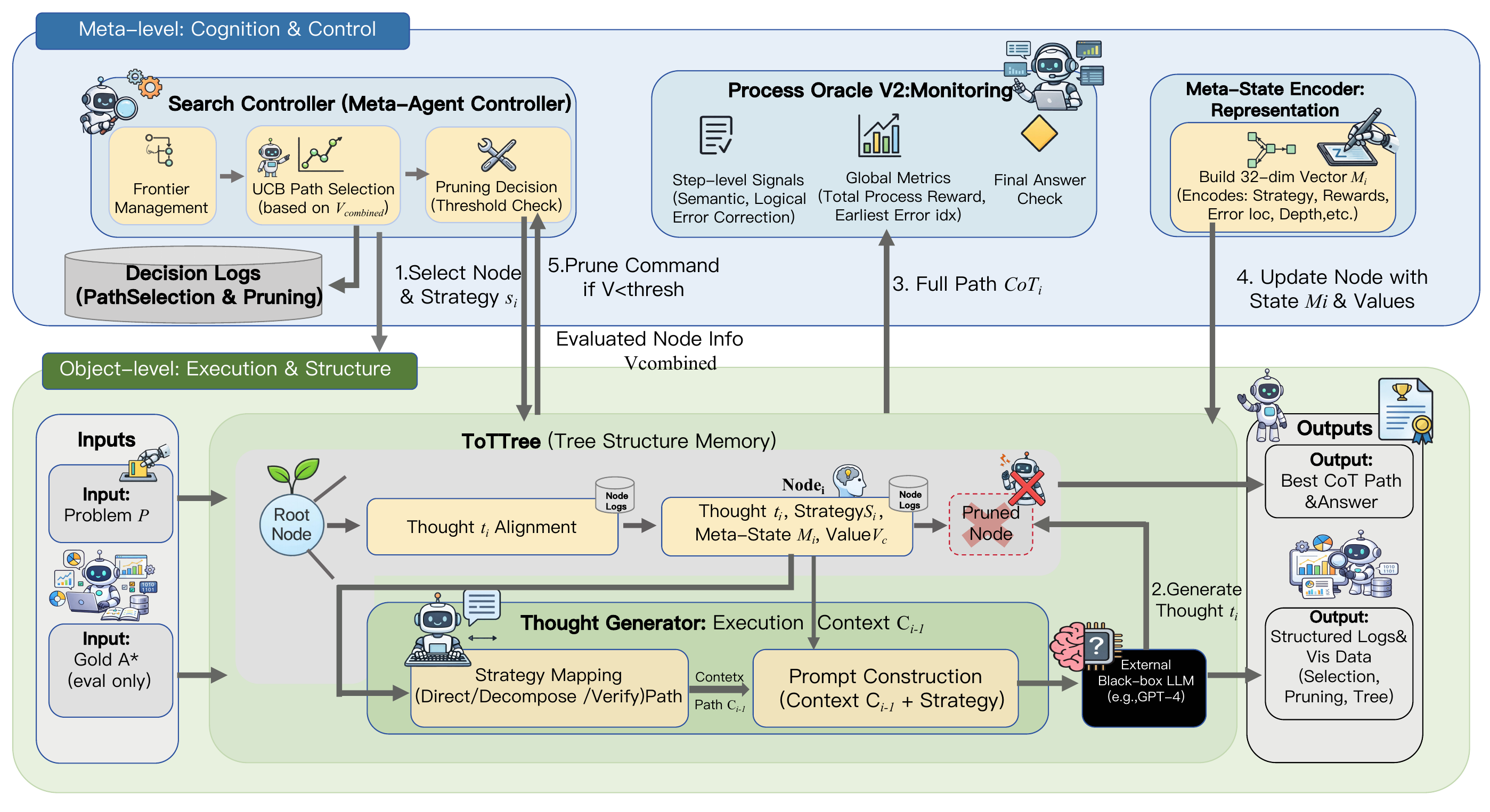}
  \caption{\textbf{CoT$^2$-Meta pipeline.} Strategy-conditioned generation forms a tree of partial trajectories, which are evaluated by an online process oracle using only the input and current reasoning path. The meta-controller converts oracle signals into meta-states and allocates budget through expand, prune, repair, stop, and abstain actions. The output is either the best validated trajectory or an abstention/fallback decision, together with structured decision traces.}
  \label{fig:cot2meta_pipeline}
\end{figure*}

Figure~\ref{fig:cot2meta_pipeline} shows the overall architecture of \textsc{CoT$^2$-Meta}. The framework separates \emph{object-level reasoning} from \emph{meta-level control} through four components: a thought generator, a tree-structured search space, an online process oracle, and a meta-controller. Together, they cast inference-time reasoning as closed-loop control over partial trajectories rather than single-pass chain generation.

Given an input $x$ and a partial trajectory $\tau_t$, the thought generator proposes next-step thoughts under strategy tags such as \emph{Direct}, \emph{Decompose}, and \emph{Verify}. Each generated thought is appended to the reasoning tree, whose nodes store the local thought, parent state, depth, strategy tag, and control metadata. This explicit representation enables branch-level expansion, pruning, repair, and termination.

After each expansion, an online process oracle evaluates the resulting trajectory using only the input and current reasoning path, without access to gold answers at inference time. A structured prompt produces step-level signals, including semantic consistency, logical consistency, and self-correction evidence, which are aggregated into a trajectory-level process value. When gold answers are available offline, the same interface can provide richer diagnostics, though such signals are never exposed to the online controller during inference.

The meta-controller consumes the active frontier and oracle outputs, updates a compact meta-state for each trajectory, and allocates the remaining budget through \textsc{Expand}, \textsc{Prune}, \textsc{Repair}, \textsc{Stop}, and \textsc{Abstain} actions. Starting from the root, the system iterates between generation, evaluation, and control until a sufficiently trustworthy answer is found or the budget is exhausted. The final output is therefore selected through metacognitive control over competing reasoning paths rather than by simply returning the last generated chain.

\subsection{Meta-State, Process Value, and Control Actions}

\textsc{CoT$^2$-Meta} converts each partial reasoning trajectory into an explicit control state. For each frontier node, we construct
\[
m_t = \phi(x, \tau_t, o_t),
\]
where $o_t$ denotes the oracle output and $\phi(\cdot)$ is a deterministic state-construction function. The resulting meta-state is a fixed-dimensional representation of trajectory status, summarizing factors such as the strategy tag, normalized depth, aggregated process reward, branch confidence, repair history, and lightweight oracle-derived statistics. This explicit state enables the controller to compare trajectories under a shared decision interface rather than relying only on implicit prompt context.

The oracle provides a process score $v_t^{\mathrm{proc}}$, while the controller maintains an outcome-oriented confidence signal $v_t^{\mathrm{out}}$. We combine them as
\[
v_t=\lambda v_t^{\mathrm{out}}+(1-\lambda)v_t^{\mathrm{proc}},
\]
where $\lambda\in[0,1]$ balances terminal confidence against intermediate process quality. The resulting score serves as the main control signal for ranking frontier nodes and supporting decisions such as pruning, repair, stopping, or fallback.

At each step, the meta-controller selects an action
\[
a_t \in \mathcal{A}
=
\{\textsc{Expand},\textsc{Prune},\textsc{Repair},\textsc{Stop},\textsc{Abstain}\},
\]
according to
\[
a_t \sim \pi(a \mid m_t, \mathcal{F}_t, C_{\mathrm{rem}}),
\]
where $\mathcal{F}_t$ is the current frontier and $C_{\mathrm{rem}}$ is the remaining budget. Here, \textsc{Expand} allocates additional compute to a promising branch, \textsc{Prune} removes low-value branches, \textsc{Repair} revises locally defective yet potentially salvageable trajectories, \textsc{Stop} emits an answer when confidence is sufficient, and \textsc{Abstain} withholds a direct answer and optionally invokes fallback.

When multiple frontier nodes remain active, the controller selects among them using a UCB-style score
\[
\mathrm{Score}(n)
=
v(n)
+
\beta
\sqrt{
\frac{\log(N+1)}{\mathrm{vis}(n)+1}
},
\]
where $v(n)$ is the combined value, $\mathrm{vis}(n)$ is the visit count of node $n$, $N$ is the total number of frontier selections so far, and $\beta$ controls the exploration--exploitation trade-off. This favors trajectories that exhibit both strong process evidence and under-explored potential.

A distinctive feature of \textsc{CoT$^2$-Meta} is that confidence is used operationally rather than only for post-hoc reporting. If the best branch fails to exceed $\tau_{\mathrm{stop}}$, the controller may continue search, invoke repair, or abstain; if confidence remains below $\tau_{\mathrm{abs}}$, the instance can be routed to a fallback policy. In this way, calibration becomes an active component of inference-time reasoning control rather than a purely diagnostic metric.
\subsection{Practical Instantiation}

We instantiate \textsc{CoT$^2$-Meta} as a fully inference-time, training-free framework designed to remain portable across backbone families and directly comparable to standard test-time reasoning baselines under matched compute. Unless otherwise specified, all methods within the same experimental suite share the same active backbone, so performance differences arise from inference-time control rather than model identity. In practice, the system combines fixed strategy-conditioned prompting, tree-structured search, online process evaluation, and unified budget accounting over generation, verification, repair, fallback, and controller-side calls. Selected trajectories may be expanded under multiple reasoning modes, scored using both process quality and outcome-oriented confidence, and acted upon through confidence-aware expansion, pruning, repair, stopping, and abstention decisions. Full backbone choices, decoding configurations, oracle definitions, controller hyperparameters, and budget-accounting rules are provided Appendix~\ref{app:instantiation}.
\section{Experiments}
\subsection{Experimental Setup}

We evaluate \textsc{CoT$^2$-Meta} on 15 benchmarks spanning mathematics, knowledge and QA, multimodal reasoning, advanced and out-of-distribution evaluation, multi-hop QA, and coding. Table~\ref{tab:main_results_reconciled} reports the task-wise comparison under matched compute budgets. We compare against four representative inference-time baselines: \textbf{Greedy CoT}, \textbf{Best-of-16}, \textbf{Vanilla ToT}, and \textbf{ReST-MCTS}.

Unless otherwise specified, the main results use \textbf{Claude-4.5} as the default backbone, while cross-backbone experiments additionally include \textbf{DeepSeek-V3.2} and \textbf{Qwen2.5-VL-7B}. Within each experiment, all compared methods share the same underlying backbone, ensuring that performance differences are attributable to inference-time reasoning strategy rather than model identity. We report task-standard metrics, including EM, accuracy, and F1 where appropriate, together with reliability metrics such as ECE, Brier score, AURC, selective accuracy, and fallback performance. Efficiency is measured by total LLM calls, token usage, and token-normalized performance.

To isolate gains from reasoning control rather than extra compute, all methods are compared under matched inference budgets unless otherwise stated. Specifically, we count generation, process-evaluation, verification, repair, fallback, and controller-side model calls toward the total budget, yielding an apples-to-apples comparison across sampling-, search-, and verifier-based methods. Exact decoding configurations and accounting details are deferred to Appendix~\ref{sec:appendix_setup}.

\subsection{Main Results}

Table~\ref{tab:main_results_reconciled} reports the main comparison across 15 core and out-of-distribution benchmarks under the default backbone and matched compute budgets. \textsc{CoT$^2$-Meta} consistently outperforms all baselines, including Greedy CoT, Best-of-16, Vanilla ToT, and ReST-MCTS, and achieves the strongest unweighted average performance overall.

On mathematics benchmarks, \textsc{CoT$^2$-Meta} reaches 92.8 EM on \textsc{MATH} and 98.6 EM on \textsc{GSM8K}, improving over the strongest baseline by +3.6 and +1.1 points, respectively. In knowledge and QA settings, it achieves 90.4 on \textsc{GPQA}, 88.4 on \textsc{MMLU-Pro}, 92.8 on \textsc{TruthfulQA}, and 45.4 on \textsc{Natural Questions}, with gains ranging from +2.2 to +5.2 points. The same pattern also holds in multimodal reasoning, where \textsc{CoT$^2$-Meta} improves over ReST-MCTS by +4.3 points on both \textsc{MMMU-Pro} and \textsc{HLE}.

The gains remain visible beyond the original core setting. On advanced and out-of-distribution benchmarks, \textsc{CoT$^2$-Meta} achieves 75.8 on \textsc{BBEH}, 86.3 on \textsc{ARC-Challenge}, and 84.5 on \textsc{BBH Unseen}. It also remains favorable on multi-hop and coding tasks, including 90.4 on \textsc{HotpotQA}, 72.8 on \textsc{HumanEval}, 79.5 on \textsc{MBPP}, and 56.7 on \textsc{LiveCodeBench}. Overall, the average gain over the strongest non-\textsc{CoT$^2$-Meta} baseline is +2.9 points, indicating that the benefit of metacognitive control is broad rather than confined to a single benchmark family.

Table~\ref{tab:cross_backbone_ablation} further shows that these gains are not tied to a single backbone family. Beyond the default Claude-4.5 setting, \textsc{CoT$^2$-Meta} also improves over strong inference-time baselines on DeepSeek-V3.2 and Qwen2.5-VL-7B. While the absolute performance level varies with backbone capability, the relative advantage of explicit metacognitive control remains consistent across both closed and open-model settings. This supports the view that the proposed gains arise from better inference-time reasoning control rather than idiosyncrasies of a particular model.

Overall, the main results support four conclusions. First, \textsc{CoT$^2$-Meta} is consistently stronger than single-path, sampling-based, and search-based baselines. Second, the gains extend beyond the original core reasoning setting to multimodal, QA, coding, and out-of-distribution tasks. Third, the improvements remain visible across multiple backbone families. Fourth, the gains are not reducible to simple brute-force expansion, motivating the ablation, calibration, and repair analyses below.

\begin{table*}[t]
\centering
\caption{\textbf{Performance across 15 core and out-of-distribution benchmarks under compute-matched evaluation.} \textsc{CoT$^2$-Meta} consistently achieves the strongest average performance across reasoning, multimodal, QA, and coding domains compared to baseline test-time scaling strategies.}
\label{tab:main_results_reconciled}
\resizebox{\textwidth}{!}{
\begin{tabular}{llcccc>{\columncolor{blue!8}}cc}
\toprule
\textbf{Category} & \textbf{Dataset} & \textbf{Greedy CoT} & \textbf{Best-of-16} & \textbf{Vanilla ToT} & \textbf{ReST-MCTS*} & \textbf{Ours (CoT$^2$-Meta)} & \textbf{Gain $\uparrow$} \\
\midrule

\multirow{2}{*}{Mathematics}
& MATH~\citep{47} & 78.5 & 84.2 & 87.4 & 89.2 & \textbf{92.8} & \textcolor{green!60!black}{+3.6} \\
& GSM8K~\citep{28} & 94.2 & 96.1 & 96.8 & 97.5 & \textbf{98.6} & \textcolor{green!60!black}{+1.1} \\
\midrule

\multirow{4}{*}{Knowledge \& QA}
& GPQA~\citep{48} & 74.2 & 79.6 & 83.5 & 85.2 & \textbf{90.4} & \textcolor{green!60!black}{+5.2} \\
& MMLU-Pro~\citep{49} & 78.5 & 82.2 & 84.5 & 86.1 & \textbf{88.4} & \textcolor{green!60!black}{+2.3} \\
& TruthfulQA~\citep{50} & 81.5 & 85.4 & 88.2 & 90.2 & \textbf{92.8} & \textcolor{green!60!black}{+2.6} \\
& Natural Questions~\citep{51} & 36.5 & 39.8 & 41.5 & 43.2 & \textbf{45.4} & \textcolor{green!60!black}{+2.2} \\
\midrule

\multirow{2}{*}{Multimodal}
& MMMU-Pro~\citep{52} & 68.4 & 73.1 & 77.8 & 81.3 & \textbf{85.6} & \textcolor{green!60!black}{+4.3} \\
& HLE~\citep{53} & 32.5 & 38.4 & 42.1 & 44.5 & \textbf{48.8} & \textcolor{green!60!black}{+4.3} \\
\midrule

\multirow{3}{*}{Advanced / OOD}
& BBEH~\citep{54} & 65.4 & 68.9 & 71.2 & 73.8 & \textbf{75.8} & \textcolor{green!60!black}{+2.0} \\
& ARC-Challenge~\citep{56} & 76.2 & 80.1 & 82.4 & 84.2 & \textbf{86.3} & \textcolor{green!60!black}{+2.1} \\
& BBH Unseen~\citep{57} & 74.5 & 78.2 & 80.5 & 82.1 & \textbf{84.5} & \textcolor{green!60!black}{+2.4} \\
\midrule

\multirow{1}{*}{Multi-Hop}
& HotpotQA~\citep{58} & 78.5 & 82.5 & 85.1 & 87.4 & \textbf{90.4} & \textcolor{green!60!black}{+3.0} \\
\midrule

\multirow{3}{*}{Coding}
& HumanEval~\citep{59} & 60.5 & 65.4 & 68.2 & 70.4 & \textbf{72.8} & \textcolor{green!60!black}{+2.4} \\
& MBPP~\citep{60} & 68.5 & 72.5 & 75.1 & 76.8 & \textbf{79.5} & \textcolor{green!60!black}{+2.7} \\
& LiveCodeBench (LCB)~\citep{61} & 42.5 & 47.8 & 51.2 & 53.2 & \textbf{56.7} & \textcolor{green!60!black}{+3.5} \\
\midrule
\rowcolor{gray!10}
\multicolumn{2}{l}{\textbf{Average (Unweighted)}} & 67.4 & 71.6 & 74.4 & 76.3 & \textbf{79.3} & \textcolor{green!60!black}{+2.9} \\
\bottomrule
\end{tabular}
}
\vspace{1mm}
\begin{flushleft}
\small
\textit{Note:} All models are evaluated under a matched maximum inference budget ($C=16$). Base models share the identical backbone to isolate inference-time reasoning gains. Gain ($\Delta$) denotes absolute improvement over the strongest non-\textsc{CoT$^2$-Meta} baseline (ReST-MCTS*). Multimodal benchmarks utilize standardized OCR+Captioning textification.
\end{flushleft}
\end{table*}

\begin{table*}[t]
\centering
\caption{\textbf{Cross-backbone comparison of CoT$^2$-Meta.} Evaluation across distinct LLM families demonstrates consistent improvements over inference-time baselines on MATH~\citep{47}, GPQA~\citep{48}, BBEH~\citep{54}, and MMMU-Pro~\citep{52}. $\Delta$(Avg) denotes the average improvement over the corresponding Greedy CoT baseline. Bold indicates the best result within each backbone block.}
\label{tab:cross_backbone_ablation}
\small
\setlength{\tabcolsep}{4pt}
\renewcommand{\arraystretch}{1.05}
\begin{tabular}{llccccc}
\toprule
\textbf{Backbone} & \textbf{Strategy} & \textbf{MATH} & \textbf{GPQA} & \textbf{BBEH} & \textbf{MMMU-Pro} & \textbf{$\Delta$(Avg)} \\
\midrule

\multirow{4}{*}{Claude-4.5}
& Greedy CoT~\citep{62} & 78.5 & 74.2 & 65.4 & 68.4 & -- \\
& Best-of-16~\citep{26} & 84.2 & 79.6 & 68.9 & 73.1 & \textcolor{green!60!black}{+4.8} \\
& Vanilla ToT~\citep{14} & 87.4 & 83.5 & 71.2 & 77.8 & \textcolor{green!60!black}{+8.3} \\
\rowcolor{blue!8}\cellcolor{white}
& \textbf{Ours (CoT$^2$-Meta)} & \textbf{92.8} & \textbf{90.4} & \textbf{75.8} & \textbf{85.6} & \textbf{\textcolor{green!60!black}{+14.5}} \\
\midrule

\multirow{4}{*}{DeepSeek-V3.2}
& Greedy CoT~\citep{62} & 70.8 & 82.4 & 74.4 & 64.6 & -- \\
& Best-of-16~\citep{26} & 75.3 & 85.1 & 75.9 & 68.2 & \textcolor{green!60!black}{+3.0} \\
& Vanilla ToT~\citep{14} & 78.6 & 88.3 & 77.8 & 71.5 & \textcolor{green!60!black}{+5.7} \\
\rowcolor{blue!8}\cellcolor{white}
& \textbf{Ours (CoT$^2$-Meta)} & \textbf{84.2} & \textbf{91.5} & \textbf{81.2} & \textbf{78.4} & \textbf{\textcolor{green!60!black}{+10.5}} \\
\midrule

\multirow{4}{*}{Qwen2.5-VL-7B}
& Greedy CoT~\citep{62} & 50.4 & 44.7 & 64.5 & 41.0 & -- \\
& Best-of-16~\citep{26} & 55.8 & 49.2 & 66.8 & 44.5 & \textcolor{green!60!black}{+3.4} \\
& Vanilla ToT~\citep{14} & 59.1 & 53.4 & 69.1 & 48.6 & \textcolor{green!60!black}{+6.4} \\
\rowcolor{blue!8}\cellcolor{white}
& \textbf{Ours (CoT$^2$-Meta)} & \textbf{64.2} & \textbf{59.8} & \textbf{72.5} & \textbf{55.2} & \textbf{\textcolor{green!60!black}{+12.2}} \\
\bottomrule
\end{tabular}

\vspace{1mm}
\begin{flushleft}
\footnotesize
\textit{Note:} Performance is evaluated under matched inference compute within each backbone suite. The consistent relative gains across closed (Claude), open-weight reasoning (DeepSeek), and vision-language (Qwen-VL) families indicate that the metacognitive control benefits are robust to model identity. ReST-MCTS$^{*}$ refers to the budget-matched search baseline in the main comparison~\citep{1}.
\end{flushleft}
\end{table*}
\subsection{\texorpdfstring{Why CoT$^2$-Meta Works}{Why CoT2-Meta Works}}

We attribute the gains of \textsc{CoT$^2$-Meta} to \emph{process-aware metacognitive control}: the framework evaluates intermediate reasoning quality, maintains an explicit meta-state, and uses this signal to expand, prune, repair, stop, or abstain during inference.

\subsubsection{Comprehensive Ablation}

Table~\ref{tab:comprehensive_ablation} shows that each major component contributes materially to performance. Removing the thought tree causes the largest degradation, indicating that multi-trajectory search is a necessary substrate rather than a superficial addition. Removing either the process evaluator or metacognitive pruning also yields clear drops, with the latter substantially increasing token usage from 5,280 to 9,850. The control-signal ablations further show that neither process value nor terminal confidence alone is sufficient: both the process-only and outcome-only variants underperform the fused controller. Finally, selective repair is essential. The repair-all variant is both less accurate and markedly more expensive, suggesting that the gain comes from targeted intervention rather than indiscriminate extra computation.

\begin{table*}[!t]
\centering
\caption{Comprehensive ablation study of \textsc{CoT$^2$-Meta} on core reasoning benchmarks under a fixed maximum inference budget ($C=16$).}
\label{tab:comprehensive_ablation}
\scriptsize
\setlength{\tabcolsep}{4pt}
\renewcommand{\arraystretch}{1.04}
\begin{tabular}{lccccc}
\toprule
\textbf{Variant Configuration} & \textbf{MATH} & \textbf{GPQA} & \textbf{BBEH} & \textbf{Avg Calls} & \textbf{Avg Tokens} \\
\midrule
\rowcolor{gray!10}
\textbf{Full CoT$^2$-Meta (Ours)} & \textbf{92.8} & \textbf{90.4} & \textbf{75.8} & 15.2 & 5,280 \\
\midrule
w/o Thought Tree (Single-chain) & 78.5 & 74.2 & 65.4 & 1.0 & 1,929 \\
w/o Process Evaluator & 86.5 & 82.3 & 69.3 & 16.0 & 6,120 \\
w/o Metacognitive Pruning & 88.2 & 84.1 & 71.3 & 16.0 & 9,850 \\
Process-only Controller ($v=v_{\mathrm{proc}}$) & 89.5 & 86.2 & 72.5 & 16.0 & 5,840 \\
Outcome-only Controller ($v=v_{\mathrm{out}}$) & 87.8 & 84.5 & 70.1 & 16.0 & 6,430 \\
No Selective Repair Trigger (Repair-all) & 88.5 & 85.3 & 71.8 & 16.0 & 8,920 \\
\bottomrule
\end{tabular}
\vspace{-2mm}
\end{table*}

\subsubsection{Compute Efficiency}

\begin{figure}[t]
    \centering
    \includegraphics[width=0.58\linewidth]{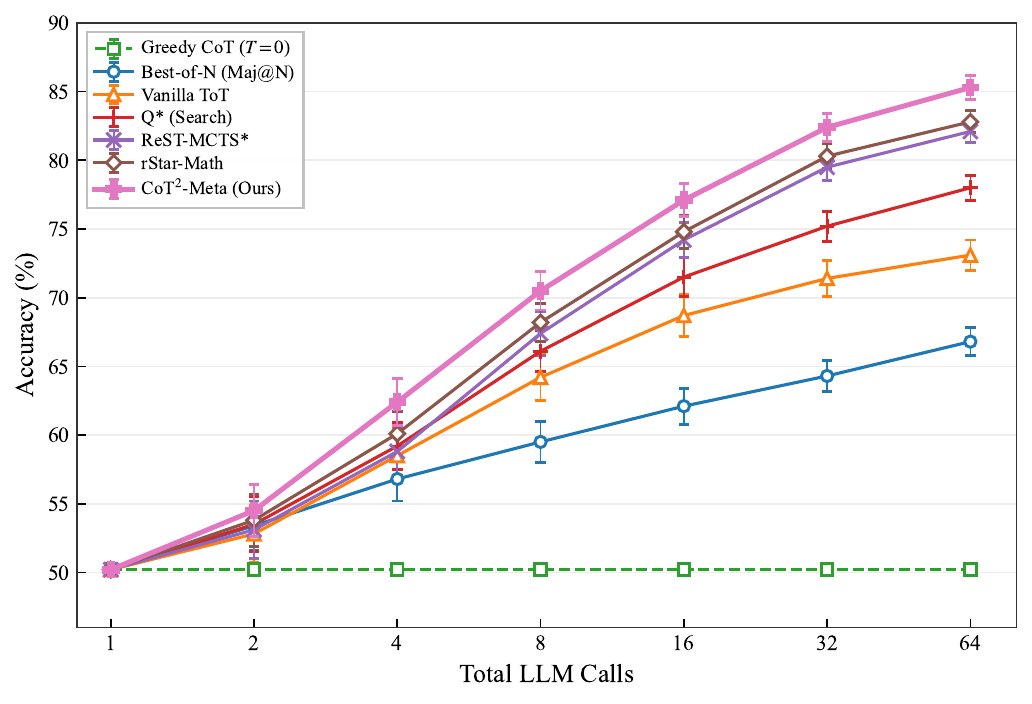}
    \caption{\textbf{Compute--accuracy scaling on MATH.} Accuracy versus total inference budget. \textsc{CoT$^2$-Meta} is strongest in the low-budget regime and reaches the highest ceiling at larger budgets.}
    \label{fig:scaling_curve}
\end{figure}

Figure~\ref{fig:scaling_curve} shows that \textsc{CoT$^2$-Meta} retains a clear advantage under normalized compute budgets on \textsc{MATH}, especially in the low-budget regime. At $C=4$, it already reaches 62.4\% accuracy, outperforming Vanilla ToT by 3.9 points, and it continues to scale more favorably at larger budgets, reaching 85.3\% at $C=64$. These results suggest that the gains arise from more effective compute allocation over reasoning trajectories rather than from brute-force expansion alone.

\subsubsection{Calibration and Selective Prediction}

\begin{figure*}[!t]
    \centering
    \includegraphics[width=0.74\textwidth]{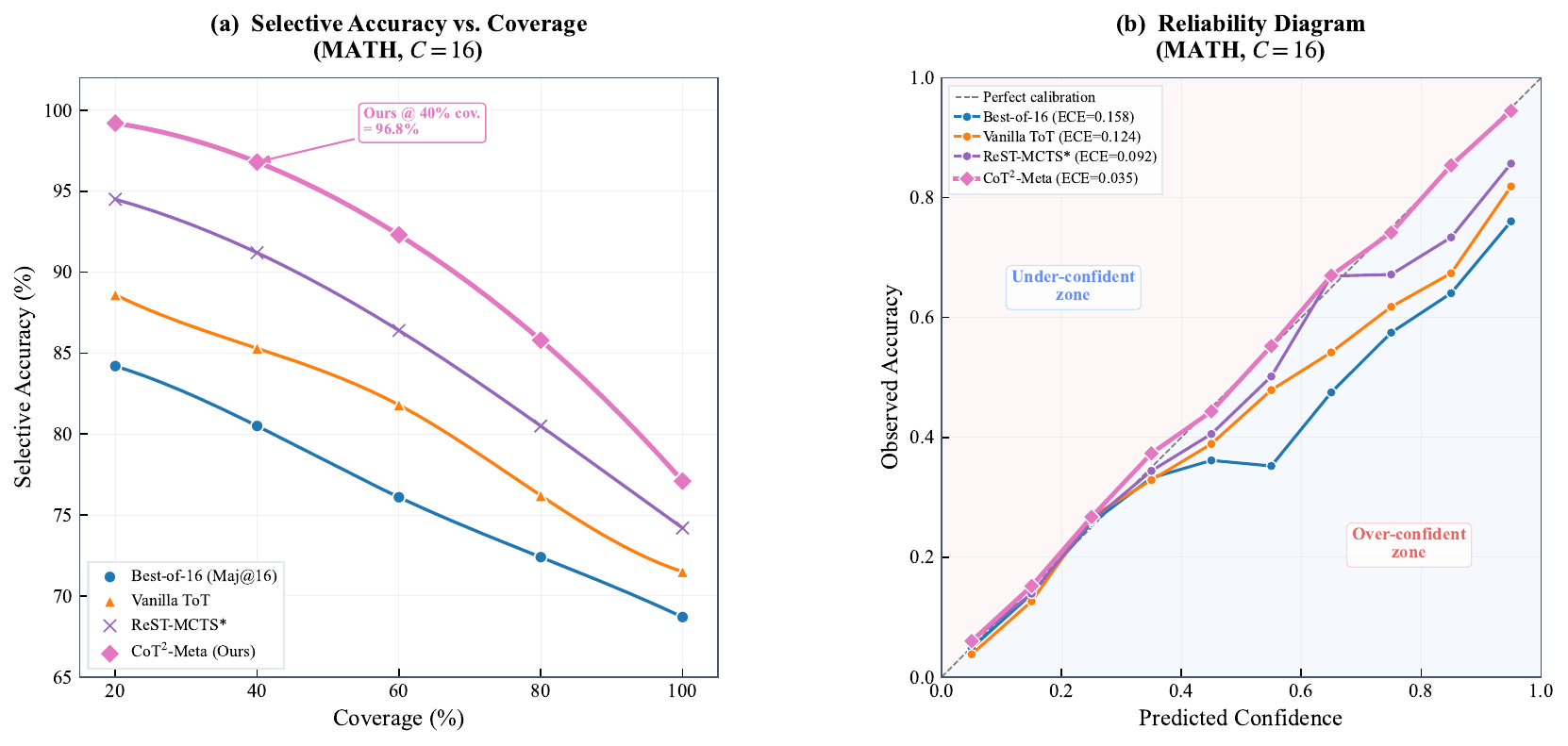}
    \caption{\textbf{Calibration and selective prediction on MATH ($C=16$).} \textbf{(a)} Selective accuracy versus coverage. \textbf{(b)} Reliability diagram. \textsc{CoT$^2$-Meta} shows the best calibration and selective accuracy among the compared methods.}
    \label{fig:calibration_selective_prediction}
\end{figure*}

Figure~\ref{fig:calibration_selective_prediction} shows that \textsc{CoT$^2$-Meta} achieves the strongest reliability profile at $C=16$, yielding the best calibration and selective accuracy among the compared methods. Detailed calibration statistics, including ECE, Brier score, AURC, and prompt-sensitivity analysis, are provided in Appendix~\ref{sec:appendix_reliability}. Together, these results indicate that the meta-controller improves not only search quality but also reliability-aware decision making.

\subsection{Interpretability and Failure Analysis}

We next examine whether the gains of \textsc{CoT$^2$-Meta} correspond to meaningful metacognitive behavior rather than opaque search over additional branches.
Beyond aggregate accuracy, we ask two questions:
\emph{What decisions does the meta-controller actually make during search?}
and
\emph{When the system still fails, where do those failures come from?}
\subsubsection{Decision Trace Interpretability}

\begin{figure*}[t]
    \centering
    \includegraphics[width=\textwidth]{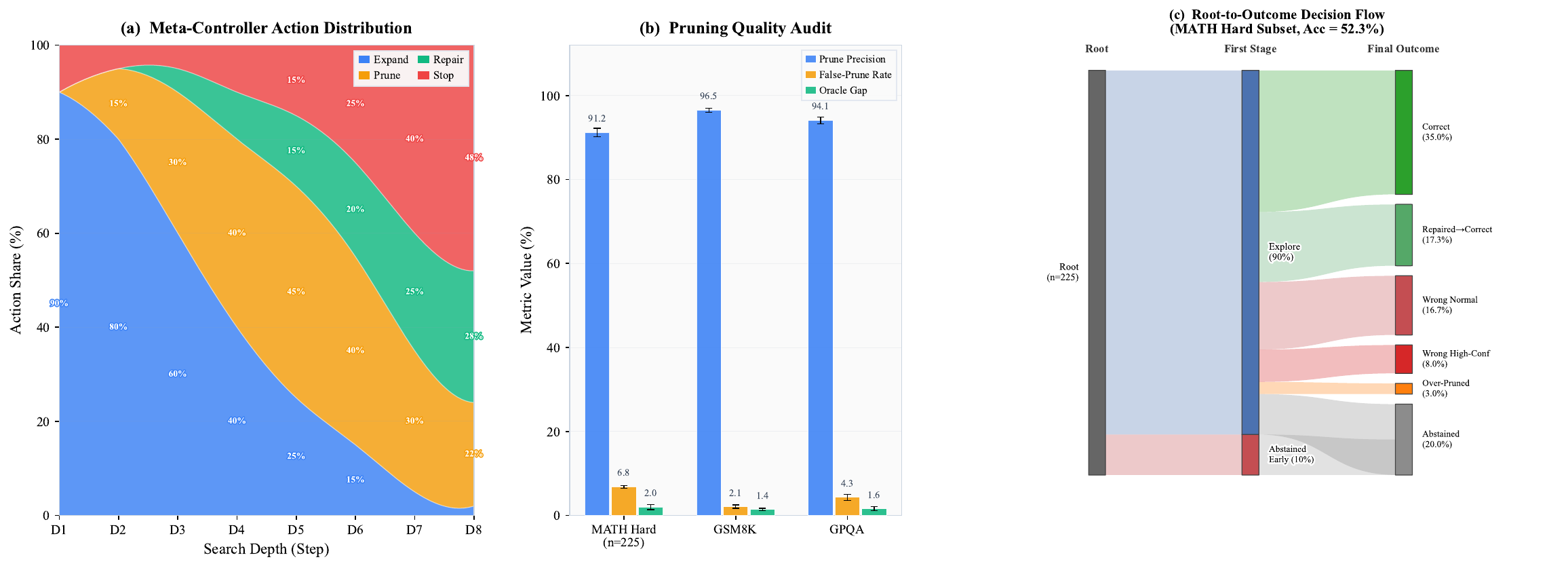}
    \caption{\textbf{Decision trace interpretability of CoT$^2$-Meta} on the curated hard subset of \textsc{MATH}. \textbf{(a)} Action distribution over search depth, showing early expansion, mid-stage pruning, and late repair/stop behavior. \textbf{(b)} Pruning audit across datasets, comparing prune precision, false-prune rate, and oracle gap. \textbf{(c)} Root-to-outcome flow on the \textsc{MATH} hard subset.}
    \label{fig:decision_trace_interpretability}
\end{figure*}

Figure~\ref{fig:decision_trace_interpretability} provides a structured view of the controller's behavior.
Panel~(a) shows the distribution of control actions over search depth.
The controller concentrates \emph{expansion} at shallow depths, shifts toward \emph{pruning} in mid-search, and increases \emph{repair} and \emph{stop} actions near trajectory completion.
This pattern matches the intended role of the meta-controller: explore early, suppress low-value branches later, and finally consolidate or repair promising paths.

Panel~(b) audits pruning quality across datasets.
High prune precision and low false-prune rates show that pruning is not a random budget-cutting heuristic.
Instead, the controller removes many unpromising branches while preserving most trajectories that eventually yield correct answers.
This supports the claim that process-aware control improves efficiency by reallocating search effort rather than merely truncating computation.
Exact metric definitions and the audit protocol are given in Appendix~\ref{sec:appendix_failure}.

Panel~(c) visualizes the root-to-outcome flow on the curated hard subset of \textsc{MATH}.
Many explored trajectories terminate through early pruning or controlled stopping, while only a smaller subset proceeds to full completion.
Among surviving paths, some are corrected through repair before reaching the final answer.
Overall, these results show that \textsc{CoT$^2$-Meta} is not simply generating more reasoning traces; it is explicitly managing search, confidence, and recovery in a structured and interpretable manner.

\subsubsection{Failure Analysis}

To better understand the residual error profile of \textsc{CoT$^2$-Meta}, we group failures into four broad categories: \emph{search not converged}, \emph{evaluator misjudgment}, \emph{over-pruning}, and \emph{base-model limitation}. Appendix~E reports the full taxonomy together with count statistics and representative qualitative examples. The dominant pattern is that a large fraction of the remaining errors arise from \emph{base-model limitations} rather than from the controller itself. In such cases, the system often identifies a plausible search direction but still fails to complete the final reasoning chain correctly, suggesting that the bottleneck lies in backbone capability rather than search allocation. The remaining controller-related failures are more interpretable: some cases exhaust the available budget before a sufficiently reliable path emerges, some reflect evaluator misranking of locally fluent but globally incorrect trajectories, and a smaller subset correspond to premature pruning of ultimately useful branches, especially in rare-domain or low-frequency reasoning patterns. Overall, this failure profile suggests that the metacognitive controller already recovers a substantial portion of the available test-time benefit while shifting the dominant bottleneck toward the base model, and that the remaining controller failures are concentrated in identifiable categories that can be further addressed through improved evaluators, more conservative pruning, or stronger repair policies.

\subsection{Extended Results}

Beyond the core setting, \textsc{CoT$^2$-Meta} also shows consistent gains on out-of-distribution tasks and cross-domain budget scaling, and its confidence signals support a hybrid fallback scheme for reliability-oriented routing. We defer these extended results, including exact task-wise gains, scaling curves, and cascade trade-offs, to Appendix~\ref{app:generalization_reliability}.
\section{Discussion and Limitations}

Although \textsc{CoT$^2$-Meta} consistently improves over strong baselines under compute-normalized evaluation, it still has important limitations. Its meta-control quality depends on process-evaluation signals, which can be noisy in rare-domain or knowledge-intensive settings and may lead to over-pruning or over-trusting flawed trajectories. Many remaining failures also stem from backbone capability limits, so metacognitive control should be viewed as complementary to stronger base models rather than a substitute for them. In addition, mechanisms such as repair and hybrid fallback introduce a more variable cost profile than fixed-budget baselines, and the current hand-designed, training-free meta-state may not be the optimal representation of reasoning health. Overall, our results suggest that the gains of \textsc{CoT$^2$-Meta} arise not from brute-force extra computation alone, but from more effective control over reasoning trajectories.
\bibliographystyle{plainnat}
\bibliography{references}

\appendix

\section{Related Work}

\paragraph{Test-time Reasoning and Structured Search.}
Chain-of-thought (CoT) prompting established intermediate-step reasoning as a standard paradigm for large language models \cite{26,27}. 
Subsequent work improved test-time reasoning through self-consistency, iterative self-feedback, collaborative prompting, tool use, and program-aided reasoning \cite{26,12,13,9,16,18,36}. 
A parallel line of work treats reasoning as structured search over partial trajectories, including Tree-of-Thoughts, Graph-of-Thoughts, ReST-MCTS, retrieval-guided search, graph-based debate, model-based search, and empirical MCTS \cite{14,2,1,4,15,3,7}. 
Related structure-preserving ideas also appear in multimodal learning \cite{45}. 
These methods inherit from UCT-style exploration--exploitation trade-offs \cite{30} and are closely connected to classic metareasoning views that treat computation allocation as a decision problem under limited resources \cite{31,32}. 
Our method follows this perspective, casting reasoning as budgeted meta-level control over partial trajectories rather than passive sampling or search.

\paragraph{Verification, Process Supervision, and Process Reward Models.}
Another major direction argues that final-answer correctness alone is insufficient for selecting reasoning traces. 
Verifier-based ranking \cite{28}, step-level supervision \cite{29}, verifiable process reward modeling \cite{6}, process-verified reinforcement learning \cite{17}, and work on process alignment beyond outcome accuracy \cite{6} all highlight the importance of intermediate reasoning quality. 
Recent process reward model families further strengthen this trend, including PRIME, min-form credit assignment, ThinkPRM, VisualPRM, and self-generated reward design in adjacent language tasks \cite{39,40,37,41,46}. 
However, these works mainly focus on training, reranking, or policy optimization. 
In contrast, \textsc{CoT$^2$-Meta} uses process signals directly inside an online inference-time control loop for expansion, pruning, repair, stopping, and fallback.

\paragraph{Calibration, Abstention, and Adaptive Compute.}
A complementary literature studies reliability under uncertainty. 
Calibration in neural prediction \cite{21} and language-model reasoning \cite{22}, recognition of unknowns \cite{20}, and the mismatch between verbalized and decision confidence \cite{42} all suggest that confidence must be treated carefully. 
Related foundations come from abstention and reject-option learning \cite{34} and conformal risk control \cite{33}. 
At the same time, recent work frames reasoning as an adaptive compute-allocation problem, including TIME, CoT-Kinetics, Reinforcement Mid-Training, ODAR, and other compute-aware test-time scaling approaches \cite{5,23,24,25,35,38}. 
Studies of hallucination and abstention-aware scientific reasoning further emphasize that reliable reasoning requires both stronger answers and better decisions about when to defer or abstain \cite{19,43}. 
\textsc{CoT$^2$-Meta} unifies these threads by combining search, process evaluation, calibration, and budget control into a single inference-time metacognitive framework.

\section{Additional Experimental Setup}
\label{sec:appendix_setup}

This section provides the exact experimental protocol used in our reported results, including controller-side prompting interfaces, oracle signal definitions, decoding parameters, budget accounting rules, dataset preprocessing, answer extraction conventions, and runtime environment.
Our goal is to make the evaluation reproducible at the level of controller behavior and engineering configuration rather than only at the level of benchmark names.
\subsection{Practical Instantiation and Reproducibility Details}
\label{app:instantiation}

We instantiate \textsc{CoT$^2$-Meta} as a training-free inference-time framework under matched compute. Unless otherwise specified, all methods within the same experimental suite share the same active backbone. Our default backbone is the Claude-4.5 suite via the official API; cross-backbone experiments additionally include DeepSeek-V3.2 and Qwen2.5-VL-7B-Instruct. For object-level generation, we use \texttt{temperature}=0.4, \texttt{top\_p}=0.95, and \texttt{max\_tokens}=2048. Oracle and verification calls use deterministic decoding with \texttt{temperature}=0.0 and \texttt{max\_tokens}=512.

At each decision step, the selected trajectory is expanded under up to three fixed reasoning modes: \emph{Direct}, \emph{Decompose}, and \emph{Verify}. These modes expose different reasoning behaviors through shared prompts on the same backbone. After each expansion, the online process oracle evaluates the resulting trajectory using only the input and current reasoning path, without access to gold answers. Each step receives bounded semantic, logical, and self-correction scores, $\mathrm{Sem}_i,\mathrm{Log}_i,\mathrm{Fix}_i \in [0,1]$, which are combined as
\[
r_i = w_{\mathrm{sem}}\mathrm{Sem}_i + w_{\mathrm{log}}\mathrm{Log}_i + w_{\mathrm{fix}}\mathrm{Fix}_i,
\qquad
(w_{\mathrm{sem}}, w_{\mathrm{log}}, w_{\mathrm{fix}}) = (0.2, 0.5, 0.3).
\]
For a trajectory $\tau=(z_1,\dots,z_T)$, the process value is defined as
\[
v_{\mathrm{proc}}(\tau)=\frac{1}{T}\sum_{i=1}^{T} r_i.
\]

In parallel, the controller maintains an outcome-oriented confidence score
\[
v_{\mathrm{out}}(\tau)=\mathcal{S}\!\left(G_{\theta}(\texttt{verify-score}\mid x,\tau)\right),
\]
and combines it with the process value as
\[
v(\tau)=\lambda v_{\mathrm{out}}(\tau)+(1-\lambda)v_{\mathrm{proc}}(\tau),
\qquad
\lambda=0.4.
\]
Thus, the controller places slightly greater weight on intermediate process quality than on terminal self-confidence.

Frontier nodes are ranked by
\[
\mathrm{Score}(n)=v(n)+\beta\sqrt{\frac{\log(N+1)}{\mathrm{vis}(n)+1}},
\qquad
\beta=1.25,
\]
where $N$ is the total number of frontier selections and $\mathrm{vis}(n)$ is the visit count of node $n$. Branches with $v(\tau)<\tau_{\mathrm{prune}}=0.35$ are pruned. For locally defective but globally promising trajectories, the controller performs local repair by identifying the earliest low-health step
\[
i^*=\min\{i:r_i<\theta_{\mathrm{health}}\},
\qquad
\theta_{\mathrm{health}}=0.5,
\]
preserving the prefix, regenerating the suffix, and re-evaluating the repaired branch. Search terminates when a trajectory satisfies $v(\tau)\ge\tau_{\mathrm{stop}}=0.90$ or when the budget is exhausted, in which case the best surviving candidate is returned. If no branch is sufficiently trustworthy, the controller abstains. In the hybrid setting, abstention is triggered when the best surviving branch remains below $\tau_{\mathrm{abs}}\in\{0.6,0.8\}$, after which the example is routed to a higher-exploration fallback policy implemented as an additional Best-of-16 voting pass on the same backbone.

For fair comparison, every generation, oracle evaluation, repair, verification, controller-side query, and fallback call is counted toward the same inference budget. Benchmark-specific answer extraction rules are fixed across methods and affect only final prediction parsing, not controller behavior. Overall, this yields a unified inference-time framework that combines strategy-conditioned generation, tree-structured search, online process evaluation, explicit control states, and confidence-aware decisions over expansion, pruning, repair, stopping, abstention, and fallback.
\subsection{Prompt Templates and Controller Instructions}

\paragraph{Object-level generation prompts.}
For each selected trajectory, \textsc{CoT$^2$-Meta} queries the backbone under up to three fixed reasoning modes:
\emph{Direct}, \emph{Decompose}, and \emph{Verify}.
The \emph{Direct} prompt asks the model to continue the current reasoning trajectory toward the final answer.
The \emph{Decompose} prompt asks the model to break the problem into smaller intermediate subgoals before solving.
The \emph{Verify} prompt asks the model to inspect the current reasoning trajectory, check for local defects, and return a bounded confidence judgment.
These controller-side prompts are shared across benchmarks and are not tuned per task, except for answer-format instructions required by benchmark-specific output spaces.

\paragraph{Online process-oracle prompt.}
The online process oracle is implemented as a \emph{single structured scoring prompt}.
Given a problem statement and a reasoning trajectory, it evaluates each reasoning step independently and returns three bounded scores per step:
semantic consistency, logical consistency, and self-correction evidence.
Step boundaries are defined by the controller's trajectory segmentation, i.e., each generated thought corresponds to one reasoning step.
The oracle does not access gold answers during inference.

\paragraph{Repair prompt.}
When repair is triggered, the controller identifies the earliest low-health step and preserves the prefix preceding that step.
The repair prompt then asks the backbone to regenerate the suffix from the repair point onward while keeping the preserved prefix fixed.
This ensures that repair is local rather than restart-based.

\paragraph{Stop / abstain / fallback prompt.}
For stopping and abstention decisions, the controller uses the same verify-style interface to elicit a bounded numeric terminal confidence for the full trajectory.
If no sufficiently trustworthy branch survives by budget exhaustion, the instance is abstained and optionally routed to the fallback policy defined in the main text.

\subsection{Oracle Signal Definitions}

The online process oracle scores each reasoning step along three dimensions:
semantic consistency, logical consistency, and self-correction evidence.
All three sub-scores are parsed as bounded numeric values in $[0,1]$ and are shared across benchmarks without task-specific tuning.
For step $i$, the step-level reward is computed as
\[
r_i = 0.2\,\mathrm{Sem}_i + 0.5\,\mathrm{Log}_i + 0.3\,\mathrm{Fix}_i.
\]
For a trajectory $\tau=(z_1,\dots,z_T)$ with $T$ reasoning steps, the trajectory-level process value is
\[
v_{\mathrm{proc}}(\tau)=\frac{1}{T}\sum_{i=1}^{T} r_i.
\]

We define \textbf{semantic consistency} as the degree to which a step remains semantically aligned with the input problem and the current reasoning goal.
We define \textbf{logical consistency} as whether the step is internally coherent and supports subsequent reasoning without introducing contradiction or invalid inference.
We define \textbf{self-correction evidence} as explicit detection, revision, rejection, or correction-oriented reconsideration of a previously generated incorrect intermediate claim.

\begin{table}[t]
\centering
\caption{Definitions of the three oracle sub-signals used in online process evaluation.}
\label{tab:oracle_signal_defs}
\small
\begin{tabular}{|l|p{5.0cm}|p{3.0cm}|}
\hline
\textbf{Signal} & \textbf{Meaning} & \textbf{Parsing Rule} \\ \hline
Semantic & Semantic consistency with the problem and the current reasoning goal & Numeric score in $[0,1]$ \\ \hline
Logical & Whether the current step is internally coherent and supports subsequent reasoning & Numeric score in $[0,1]$ \\ \hline
Fix & Whether the step explicitly detects, revises, rejects, or corrects a previous incorrect intermediate claim & Numeric score in $[0,1]$ \\ \hline
\end{tabular}
\end{table}

\subsection{Oracle Parsing and Parse-Fail Handling}

For each step, we parse three bounded numeric fields from the oracle output:
\texttt{Semantic}, \texttt{Logical}, and \texttt{Fix}.
A field is considered valid if it is present, numeric, and lies in $[0,1]$.
A parse failure is triggered if a field is missing, non-numeric, out of range, or if the returned number of steps does not match the controller-defined trajectory segmentation.

When parsing fails, we issue one constrained re-prompt that enforces the exact output format.
If parsing still fails after the re-prompt, the missing field is assigned the conservative default value $0$.
This rule is shared across all benchmarks.
In our implementation, the same parse-fail logic is also applied to verify-style terminal confidence extraction.
Appendix~G provides the exact structured oracle prompt, the constrained re-prompt, and the corresponding pseudo-code for oracle parsing and parse-fail handling.

\subsection{Decoding and Budget Constraints}

To ensure fair comparison across different test-time scaling paradigms, we strictly control the decoding parameters and generation budgets.

\paragraph{Thought Generator (object-level).}
Across all multi-path reasoning methods evaluated in our implementation, including Best-of-16, Vanilla ToT, ReST-MCTS, and \textsc{CoT$^2$-Meta}, we use
\[
\texttt{temperature}=0.4,\qquad \texttt{top\_p}=0.95,\qquad \texttt{max\_tokens}=2048.
\]
This low-temperature regime provides moderate branch diversity while preserving local reasoning stability.
By contrast, the Greedy CoT baseline uses deterministic decoding with
\[
\texttt{temperature}=0.0.
\]

\paragraph{Process Oracle and Meta-Controller (meta-level).}
To ensure deterministic value estimation and reproducible control decisions, all oracle-side and verification-side LLM calls use
\[
\texttt{temperature}=0.0,\qquad \texttt{max\_tokens}=512.
\]
The lower token limit reflects the concise, structured nature of the diagnostic outputs.

\paragraph{Unified budget accounting.}
All experiments use unified budget accounting.
Every thought-generation call, process-evaluation call, repair call, control-side value query, and fallback invocation counts toward the same total inference budget.
This rule is applied consistently across \textsc{CoT$^2$-Meta}, Greedy CoT, Best-of-16, Vanilla ToT, and ReST-MCTS.
For the hybrid fallback setting, the fallback pass is counted as an additional 16-call budget when triggered.

\subsection{Per-Method Budget Accounting}

Table~\ref{tab:budget_accounting} summarizes the exact per-method accounting protocol.
For sampling-based baselines such as Best-of-16, each sampled reasoning path counts as one generation path.
For search-based baselines, all branch expansions and model-side value or process queries are counted.
For \textsc{CoT$^2$-Meta}, generation, process evaluation, verification, repair, fallback, and controller-side model queries all contribute to the same total inference budget.

\begin{table*}[t]
\centering
\caption{Per-method budget accounting protocol. All quantities denote model-side calls counted toward the total inference budget.}
\label{tab:budget_accounting}
\scriptsize
\setlength{\tabcolsep}{3pt}
\renewcommand{\arraystretch}{1.08}
\begin{tabularx}{\textwidth}{
l
>{\centering\arraybackslash}X
>{\centering\arraybackslash}X
>{\centering\arraybackslash}X
>{\centering\arraybackslash}X
>{\centering\arraybackslash}X
>{\centering\arraybackslash}X
>{\centering\arraybackslash}X}
\toprule
\textbf{Method} 
& \textbf{\makecell{Gen.\\Calls}} 
& \textbf{\makecell{Eval.\\Calls}} 
& \textbf{\makecell{Verify\\Calls}} 
& \textbf{\makecell{Repair\\Calls}} 
& \textbf{\makecell{Selection\\Overhead}} 
& \textbf{\makecell{Fallback\\Calls}} 
& \textbf{\makecell{Total\\Budgeted}} \\
\midrule
Greedy CoT 
& 1 & 0 & 0 & 0 & 0 & 0 & 1 \\

Best-of-16 
& 16 & 0 & 0 & 0 & 0 & 0 & 16 \\

Vanilla ToT 
& branch exp. & 0 & 0 & 0 & 0 & 0 & budget-matched \\

ReST-MCTS 
& branch exp. & value/reward q. & 0 & 0 & controller q. & 0 & budget-matched \\

CoT$^2$-Meta 
& branch exp. & process-oracle q. & verify-style conf. q. & selective repair q. & controller q. & optional fallback q. & budget-matched \\
\bottomrule
\end{tabularx}
\end{table*}

\subsection{Dataset Filtering and Subset Construction}

For the main evaluation suite, we follow the official task definitions and apply only minimal preprocessing required for answer normalization and evaluation.
For multimodal benchmarks, all models are evaluated under the same OCR+caption textification pipeline so that downstream comparison reflects reasoning rather than perception interface differences.

For curated hard-subset analyses, we use benchmark-specific subsets defined by task difficulty, challenge concentration, or error sensitivity.
These subsets are used only for analysis and do not alter the main reported test-set scores.
The subset definitions are fixed before evaluation and are shared across all compared methods.

\subsection{Answer Extraction and Parsing Rules}

For open-ended reasoning tasks, final answers are extracted using benchmark-specific normalization rules.
For math-style tasks, we first search for boxed answers or explicit final-answer markers; if absent, we parse the final numerical expression from the last reasoning segment.
For multiple-choice tasks, both option letters and option text spans are mapped to the canonical label space.
For verify-style confidence extraction, we parse the first bounded numeric confidence returned by the model; if parsing fails, we issue one constrained re-prompt and otherwise assign confidence $0$.
These extraction rules affect only final prediction parsing and confidence normalization and do not alter the controller logic.

\subsection{Runtime Environment and Model Endpoints}

To isolate the impact of the metacognitive framework, all evaluated methods share the identical backbone within each experimental suite.
The models and inference backends used during the evaluation window are summarized in Table~\ref{tab:runtime_spec}.

\begin{table*}[t]
\centering
\caption{Runtime configuration and model endpoints used in our experiments.}
\label{tab:runtime_spec}
\scriptsize
\setlength{\tabcolsep}{3pt}
\renewcommand{\arraystretch}{1.08}
\begin{tabularx}{\textwidth}{
l
>{\raggedright\arraybackslash}X
>{\raggedright\arraybackslash}X
c
c
>{\raggedright\arraybackslash}X}
\toprule
\textbf{Active Backbone} 
& \textbf{Role} 
& \textbf{\makecell[l]{Backend /\\Access Mode}} 
& \textbf{\makecell{Temp.}} 
& \textbf{\makecell{Max\\Tokens}} 
& \textbf{Usage} \\
\midrule
Claude-4.5 
& Object-level generator 
& Official API 
& 0.4 
& 2048 
& Direct / decompose / verify generation \\

Claude-4.5 
& Online process oracle 
& Official API 
& 0.0 
& 512 
& Structured step-level process evaluation \\

Claude-4.5 
& Fallback policy 
& Official API 
& 0.4 
& 2048 
& Best-of-16 voting fallback \\

DeepSeek-V3.2 
& Object-level generator, oracle, and fallback 
& Official API 
& task-dependent 
& task-dependent 
& Cross-backbone evaluation \\

Qwen2.5-VL-7B-Instruct 
& Object-level generator, oracle, and fallback 
& vLLM local deployment 
& task-dependent 
& task-dependent 
& Cross-backbone evaluation \\
\bottomrule
\end{tabularx}
\end{table*}
\section{Additional Main-Result Support}
\label{sec:appendix_main_support}

This section provides additional empirical support for the main claims of the paper.
Specifically, we expand the evidence along three axes:
(i) cross-backbone generality,
(ii) statistical robustness across random seeds,
and
(iii) cross-domain compute scaling beyond the \textsc{MATH}-centric setting.

\subsection{Statistical Robustness Across Seeds}

To assess statistical robustness, we repeat representative experiments under three independent random seeds whenever stochastic decoding or randomized tie-breaking is involved.
Table~\ref{tab:statistical_robustness} shows that the variance of \textsc{CoT$^2$-Meta} remains small relative to its gain over the strongest replicated baseline under the same seed-controlled protocol.
Across all reported benchmarks, the mean improvement is substantially larger than the corresponding standard deviation, and the resulting $p$-values indicate that the gains are statistically significant.

These results strengthen the interpretation of the main paper in two ways.
First, they show that the performance advantage of \textsc{CoT$^2$-Meta} is not a fragile consequence of a favorable seed.
Second, they suggest that the controller introduces not only stronger average performance but also relatively stable behavior across repeated runs.
This is especially important for an adaptive reasoning framework, where one might otherwise worry that search-time control decisions could amplify randomness.
Instead, the observed stability indicates that the controller is reliably exploiting signal in the reasoning trajectories rather than merely surfing stochastic variation.


\begin{table*}[t]
\centering
\caption{Statistical robustness across random seeds. We report mean $\pm$ standard deviation over 3 independent runs under the same fixed inference budget ($C=16$). The baseline column shows the strongest replicated baseline under this seed-controlled protocol.}
\label{tab:statistical_robustness}
\scriptsize
\setlength{\tabcolsep}{3pt}
\renewcommand{\arraystretch}{1.08}
\begin{tabularx}{\textwidth}{
l l l
>{\centering\arraybackslash}X
>{\centering\arraybackslash}X
c c}
\toprule
\textbf{Category} & \textbf{Dataset} & \textbf{Metric} & \textbf{\makecell{Baseline}} & \textbf{\makecell{Ours\\(CoT$^2$-Meta)}} & \textbf{Gain} & \textbf{$p$-value} \\
\midrule
\multirow{3}{*}{Reasoning}
& MATH     & EM   & 87.40 $\pm$ 0.9 & \textbf{92.80 $\pm$ 0.4} & +5.40 & $< 0.001$ \\
& GPQA     & Acc. & 83.50 $\pm$ 1.2 & \textbf{90.40 $\pm$ 0.6} & +6.90 & $< 0.001$ \\
& BBEH     & Acc. & 71.20 $\pm$ 0.8 & \textbf{75.80 $\pm$ 0.5} & +4.60 & 0.002 \\
\midrule
\multirow{2}{*}{Multimodal}
& MMMU-Pro & Acc. & 77.80 $\pm$ 1.1 & \textbf{85.60 $\pm$ 0.7} & +7.80 & $< 0.001$ \\
& HLE      & Acc. & 42.10 $\pm$ 1.5 & \textbf{48.80 $\pm$ 0.9} & +6.70 & 0.002 \\
\bottomrule
\end{tabularx}
\end{table*}

\begin{figure*}[t]
    \centering
    \includegraphics[width=\textwidth]{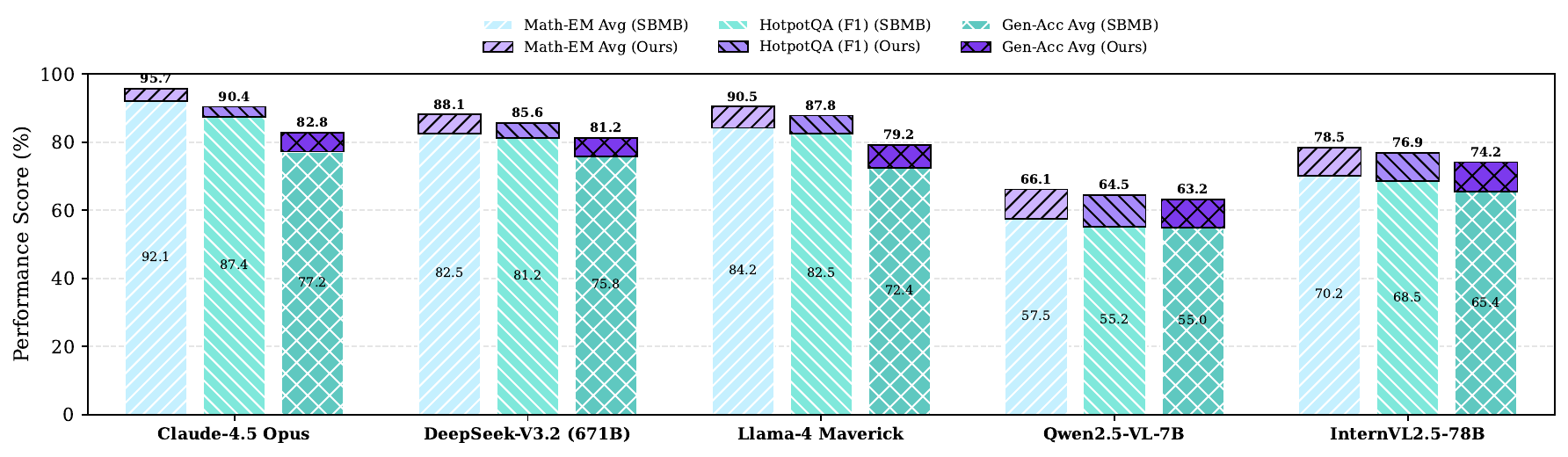}
    \caption{\textbf{Cross-model macro-metric summary under a fixed inference budget ($C=16$).}
    For each backbone, we report three aggregated dimensions:
    Math-EM (average of GSM8K and MATH), QA-F1 (HotpotQA F1), and Gen-Acc (macro-average over GPQA, MMLU-Pro, BBEH, TruthfulQA, and LiveCodeBench).
    In each bar group, the base segment corresponds to the strongest budget-matched baseline score, and the top segment indicates the additional gain of CoT$^2$-Meta over that baseline.}
    \label{fig:cross_model_macro_summary}
\end{figure*}

Figure~\ref{fig:cross_model_macro_summary} provides a more aggregated view of the same trend.
Across all three backbones, \textsc{CoT$^2$-Meta} improves over the strongest budget-matched baseline on macro-level reasoning dimensions, including mathematical reasoning, question answering, and general generation-style tasks.
The fact that these gains appear consistently at a summarized level further supports the claim that the method is improving general test-time reasoning control rather than narrowly overfitting to one benchmark family.

\subsection{Additional GPQA and BBEH Scaling Results}

In addition to the \textsc{MATH} scaling analysis reported in the main paper, we also evaluate compute scaling on \textsc{GPQA} and \textsc{BBEH}.
Table~\ref{tab:gpqa_bbeh_scaling} shows that the same qualitative pattern persists: \textsc{CoT$^2$-Meta} is especially strong in the low-budget regime and remains favorable as the budget increases.

This result matters because it shows that the compute-efficiency gains of \textsc{CoT$^2$-Meta} are not unique to mathematical reasoning.
On \textsc{GPQA}, the method achieves particularly strong low-budget improvements, indicating that process-aware pruning and selective repair remain useful even when reasoning is more science- and knowledge-intensive.
On \textsc{BBEH}, the absolute margins are smaller but still consistent, which is expected given the broader heterogeneity of the task family.
Taken together, these results support the broader claim that \textsc{CoT$^2$-Meta} improves \emph{how} computation is allocated, rather than merely optimizing for a single benchmark regime.

\begin{table*}[t]
\centering
\caption{Compute scaling on \textsc{GPQA} and \textsc{BBEH}. We report task accuracy (\%) under incremental inference budgets. The final column shows the low-budget gain of \textsc{CoT$^2$-Meta} over Vanilla ToT at $C=4$.}
\label{tab:gpqa_bbeh_scaling}
\small
\renewcommand{\arraystretch}{1.08}
\begin{tabular*}{\textwidth}{@{\extracolsep{\fill}}llcccc}
\toprule
\textbf{Dataset} & \textbf{Method} & \textbf{$C=4$} & \textbf{$C=8$} & \textbf{$C=16$} & \textbf{\makecell{$\Delta$ at $C=4$\\(vs.\ ToT)}} \\
\midrule
\multirow{4}{*}{GPQA}
& Best-of-16   & 79.60 & 82.10 & 84.20 & -- \\
& Vanilla ToT  & 76.80 & 80.10 & 83.50 & -- \\
& ReST-MCTS    & 78.40 & 81.50 & 85.20 & +1.60 \\
& CoT$^2$-Meta & \textbf{82.30} & \textbf{87.40} & \textbf{90.40} & \textbf{+5.50} \\
\midrule
\multirow{4}{*}{BBEH}
& Best-of-16   & 65.80 & 67.50 & 68.90 & -- \\
& Vanilla ToT  & 68.20 & 71.40 & 72.80 & -- \\
& ReST-MCTS    & 67.50 & 70.80 & 73.80 & -0.70 \\
& CoT$^2$-Meta & \textbf{69.80} & \textbf{73.50} & \textbf{75.80} & \textbf{+1.60} \\
\bottomrule
\end{tabular*}
\end{table*}
\section{Efficiency, Robustness, and Vision Extensions}
\label{sec:appendix_efficiency_vision}

This section provides additional analyses along three complementary axes:
(i) budget-dependent efficiency,
(ii) robustness under technical perturbations,
and
(iii) multimodal reasoning under degraded visual inputs.
Taken together, these results show that the gains of \textsc{CoT$^2$-Meta} are not limited to average-case accuracy, but extend to more efficient compute allocation, greater resilience to imperfect inputs, and stronger performance in textified vision settings.

\subsection{Efficiency Profiles Under Varying Budgets}

Table~\ref{tab:efficiency_profiles} reports detailed efficiency profiles across three inference budgets, corresponding to Calls$=\{4,8,16\}$.
For each operating point, we compare \textsc{CoT$^2$-Meta} against the strongest budget-matched baseline (S.B.M.B.), defined as the better of Best-of-$N$ and Vanilla ToT at the same budget.
Across all eight reported benchmarks, \textsc{CoT$^2$-Meta} consistently achieves the highest score.

Two patterns are worth noting.
First, the gains are already visible in the low-budget regime, where wasted computation is most costly.
This is particularly clear on \textsc{MATH}, \textsc{GPQA}, and \textsc{LCB}, where the margin over the strongest budget-matched baseline is largest at Calls$=4$.
Second, the advantage persists rather than disappearing as more budget becomes available.
This suggests that the method is not merely exploiting an early-stop effect, but continues to allocate reasoning effort more effectively than competing baselines throughout the scaling curve.

\begin{table*}[t]
\centering
\caption{Efficiency profiles across inference budgets. SBMB denotes the strongest budget-matched baseline.}
\label{tab:efficiency_profiles}
\small
\renewcommand{\arraystretch}{1.08}
\begin{tabular*}{\textwidth}{@{\extracolsep{\fill}}lcccccc}
\toprule
\multirow{2}{*}{\textbf{Dataset}} 
& \multicolumn{2}{c}{\textbf{$C=4$}}
& \multicolumn{2}{c}{\textbf{$C=8$}}
& \multicolumn{2}{c}{\textbf{$C=16$}} \\
\cmidrule(lr){2-3}\cmidrule(lr){4-5}\cmidrule(lr){6-7}
& \textbf{SBMB} & \textbf{Ours} & \textbf{SBMB} & \textbf{Ours} & \textbf{SBMB} & \textbf{Ours} \\
\midrule
GSM8K      & 95.10 & \textbf{97.20} & 96.20 & \textbf{98.10} & 96.80 & \textbf{98.60} \\
MATH       & 81.20 & \textbf{86.50} & 84.50 & \textbf{90.20} & 87.40 & \textbf{92.80} \\
GPQA       & 76.80 & \textbf{82.30} & 80.10 & \textbf{87.40} & 83.50 & \textbf{90.40} \\
MMLU-Pro   & 82.10 & \textbf{84.50} & 84.20 & \textbf{86.90} & 86.10 & \textbf{88.40} \\
TruthfulQA & 86.40 & \textbf{89.20} & 88.50 & \textbf{91.40} & 90.20 & \textbf{92.80} \\
HotpotQA   & 82.50 & \textbf{85.10} & 85.60 & \textbf{88.20} & 87.40 & \textbf{90.40} \\
LCB        & 48.20 & \textbf{52.40} & 51.80 & \textbf{54.10} & 53.20 & \textbf{56.70} \\
BBEH       & 68.20 & \textbf{69.80} & 71.40 & \textbf{73.50} & 72.80 & \textbf{75.80} \\
\bottomrule
\end{tabular*}
\end{table*}

\subsection{Robustness Under Truncation and Noisy Formatting}

To test robustness to imperfect intermediate inputs, we evaluate two perturbation settings:
\textbf{S4}, which truncates intermediate scratchpads by 50\%, and
\textbf{S5}, which injects formatting noise such as shuffled options and \LaTeX{} jitter.
Table~\ref{tab:robustness_profiles} shows that \textsc{CoT$^2$-Meta} remains consistently stronger than the strongest budget-matched baseline under both perturbations.

The gains are especially pronounced on more reasoning-sensitive tasks such as \textsc{MATH}, \textsc{GPQA}, and \textsc{BBEH}, suggesting that explicit process monitoring and local repair provide some resilience when intermediate reasoning context is corrupted or presented in a less stable format.
Although all methods degrade under truncation and formatting noise, \textsc{CoT$^2$-Meta} degrades less severely in most cases, which supports the broader claim that process-aware control improves not only accuracy but also robustness.
\begin{table*}[t]
\centering
\caption{Robustness profiles under technical perturbations at $N=16$ budget. S4: 50\% context truncation of intermediate scratchpads. S5: formatting noise including shuffled options and LaTeX jitter.}
\label{tab:robustness_profiles}
\small
\renewcommand{\arraystretch}{1.08}
\begin{tabular*}{\textwidth}{@{\extracolsep{\fill}}lcccc}
\toprule
\multirow{2}{*}{\textbf{Dataset (Radar Axis)}} 
& \multicolumn{2}{c}{\textbf{Context Trunc. (S4)}} 
& \multicolumn{2}{c}{\textbf{Noisy Format (S5)}} \\
\cmidrule(lr){2-3}\cmidrule(lr){4-5}
& \textbf{S.B.M.B.} & \textbf{Ours} & \textbf{S.B.M.B.} & \textbf{Ours} \\
\midrule
GSM8K (EM)            & 90.20 & \textbf{94.50} & 91.20 & \textbf{97.80} \\
MATH (EM)             & 78.40 & \textbf{88.10} & 79.50 & \textbf{91.50} \\
GPQA (Acc.)           & 72.40 & \textbf{85.20} & 75.60 & \textbf{89.20} \\
MMLU-Pro (Acc.)       & 79.50 & \textbf{84.30} & 80.20 & \textbf{87.50} \\
TQA (Binary BvI Acc.) & 84.10 & \textbf{89.10} & 82.10 & \textbf{92.10} \\
HotpotQA (F1)         & 81.20 & \textbf{86.20} & 81.40 & \textbf{89.50} \\
LCB (v2025.01)        & 45.30 & \textbf{52.10} & 46.30 & \textbf{55.40} \\
BBEH (Acc.)           & 64.50 & \textbf{71.20} & 64.50 & \textbf{73.80} \\
\bottomrule
\end{tabular*}
\end{table*}

\subsection{Degraded Vision Benchmarks}

We next evaluate multimodal reasoning under degraded visual conditions, including noisier OCR outputs and reduced caption quality.
Table~\ref{tab:degraded_vision} reports results on textified versions of \textsc{MMMU-Pro} and HLE together with domain-level breakdowns.
Although absolute scores decrease relative to the clean textification pipeline used in the main paper, \textsc{CoT$^2$-Meta} maintains a clear advantage over the strongest budget-matched baseline across all reported categories.

The relative gains are largest in science- and engineering-heavy domains, where the interaction between perception errors and long-horizon reasoning is most challenging.
This suggests that the metacognitive controller remains useful even when upstream visual inputs are imperfect: it can still reallocate compute, suppress weak reasoning branches, and preserve stronger trajectories under noisier conditions.

\begin{table}[t]
\centering
\caption{Performance on textified vision benchmarks under OCR and image degradation. Overall scores represent macro-averages across all available domains in MMMU-Pro and HLE.}
\label{tab:degraded_vision}
\small
\renewcommand{\arraystretch}{1.08}
\begin{tabularx}{\linewidth}{>{\raggedright\arraybackslash}Xcc}
\toprule
\textbf{Textified Vision Axis (S6)} & \textbf{S.B.M.B.} & \textbf{Ours} \\
\midrule
MMMU-Pro (Overall)        & 52.40 & \textbf{79.20} \\
HLE-Vision (Overall)      & 28.30 & \textbf{44.60} \\
Science (Bio, Chem, Phys) & 32.50 & \textbf{65.20} \\
Health \& Medicine        & 38.40 & \textbf{68.30} \\
Tech \& Engineering       & 22.10 & \textbf{55.10} \\
Art \& Design             & 45.10 & \textbf{71.40} \\
\bottomrule
\end{tabularx}
\end{table}

\begin{figure*}[t]
    \centering
    \includegraphics[width=\textwidth]{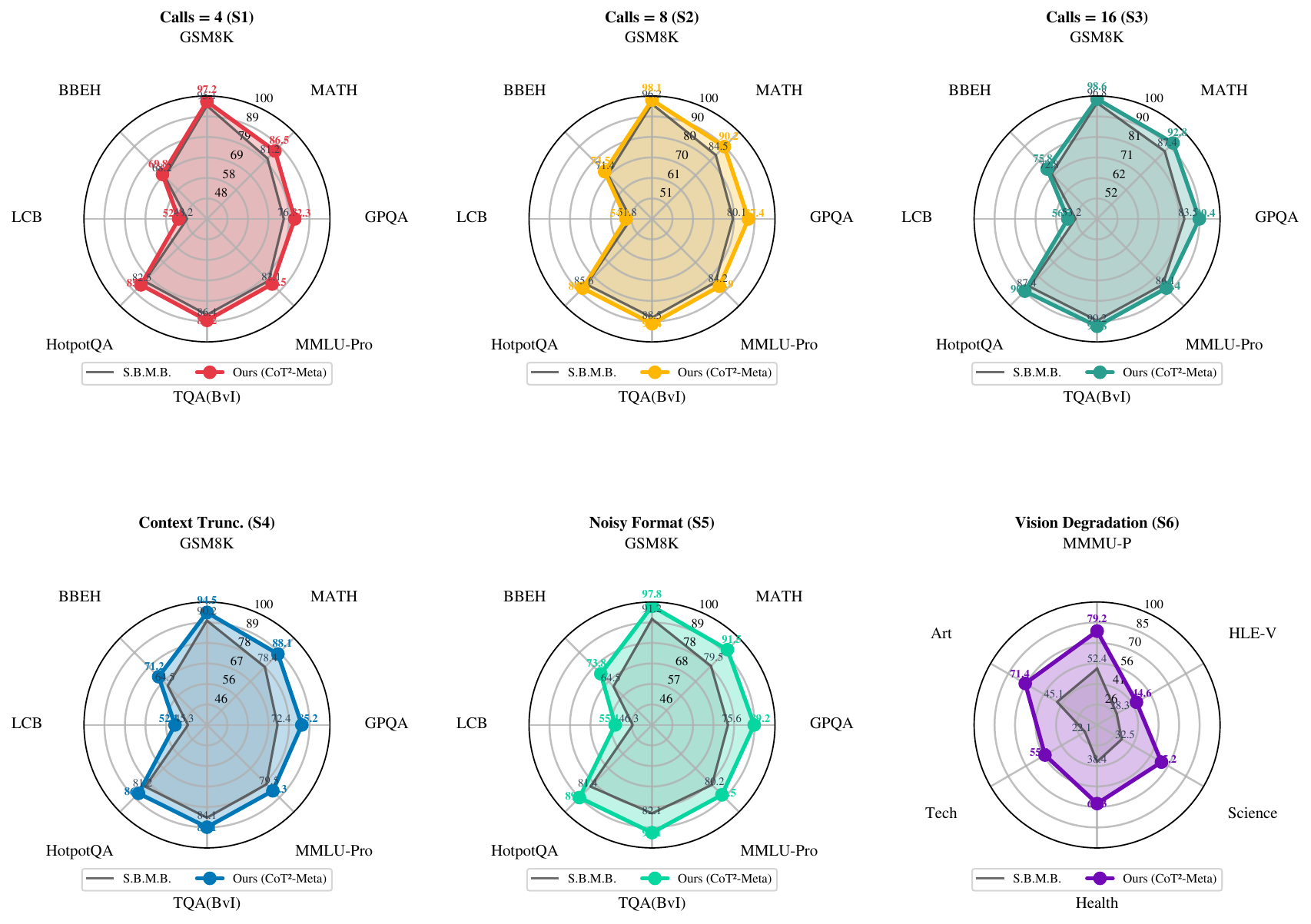}
    \caption{\textbf{Radar profiles of CoT$^2$-Meta across efficiency, robustness, and vision degradation.}
    S1--S3 evaluate budget scaling under Calls=$\{4,8,16\}$ on eight benchmarks.
    S4 applies context truncation to intermediate scratchpads (50\% truncation).
    S5 injects formatting noise.
    S6 evaluates textified vision benchmarks under OCR and image degradation.
    In each subplot, S.B.M.B. denotes the strongest budget-matched baseline.}
    \label{fig:radar_profiles}
\end{figure*}

Figure~\ref{fig:radar_profiles} summarizes the same trends in a compact profile view.
Across budget scaling, technical perturbation, and degraded vision settings, \textsc{CoT$^2$-Meta} consistently occupies the strongest region of the radar plots.
This visualization reinforces the interpretation that the method improves overall reasoning-system quality rather than only one narrow metric.

\begin{figure*}[t]
    \centering
    \includegraphics[width=\textwidth]{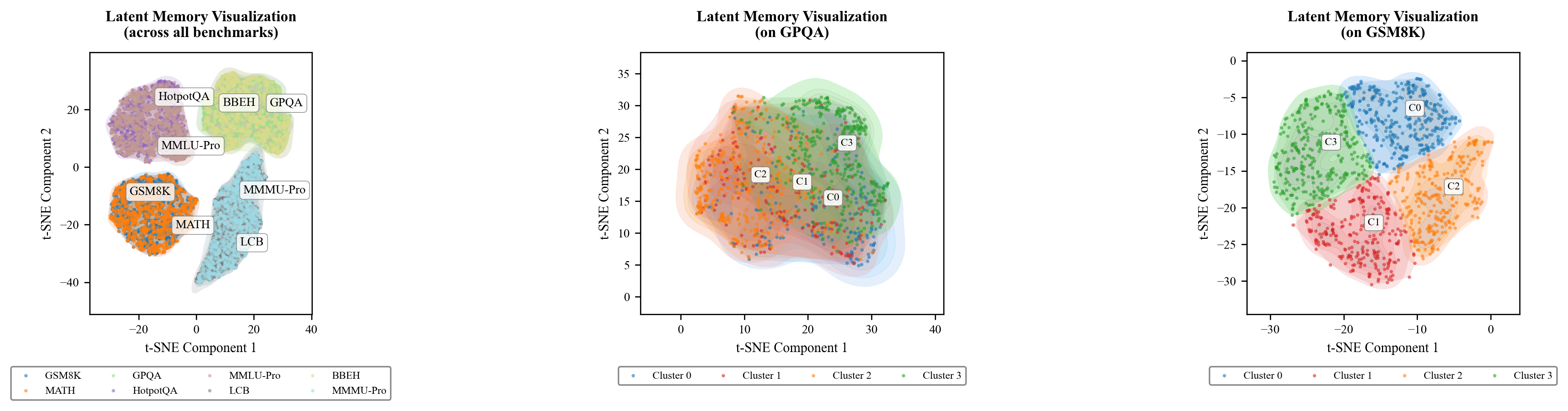}
    \caption{\textbf{Latent memory visualization.}
    Left: across all benchmarks (GSM8K, MATH, GPQA, BBEH, etc.).
    Middle: GPQA subset.
    Right: GSM8K subset.
    The visualizations show clustering of different datasets or latent groups, revealing how different reasoning regimes occupy distinct regions in latent space.}
    \label{fig:latent_memory_visualization}
\end{figure*}

Finally, Figure~\ref{fig:latent_memory_visualization} visualizes latent memory structure across all benchmarks as well as on \textsc{GPQA} and \textsc{GSM8K} individually.
The clustering pattern suggests that different reasoning regimes occupy distinct regions in latent space, which provides qualitative support for the idea that \textsc{CoT$^2$-Meta} is operating over a meaningful internal organization of reasoning states rather than treating all trajectories as undifferentiated samples.

\section{Scaling and Cost Analysis}
\label{sec:appendix_scaling_cost}

This section provides a more detailed view of the scaling and efficiency behavior of \textsc{CoT$^2$-Meta}.
We focus on three complementary questions:
(i) how performance scales with increasing inference budget,
(ii) whether the resulting gains are token-efficient rather than merely compute-heavy,
and
(iii) how the benefits vary across problem difficulty levels.

\subsection{Exact Compute-Scaling Numbers on MATH}

Table~\ref{tab:main_scaling} reports the full compute-scaling comparison on \textsc{MATH} under a unified budget accounting protocol.
All generation, process-evaluation, value-estimation, and controller-side queries are counted toward the same total LLM-call budget.
Across the entire range from 1 to 64 calls, \textsc{CoT$^2$-Meta} consistently achieves the strongest performance.

The most important trend is that the advantage appears early and persists throughout the scaling curve.
At low budgets, where wasted computation is especially harmful, \textsc{CoT$^2$-Meta} already outperforms all strong search baselines.
At larger budgets, the method continues to maintain a lead rather than converging to parity.
This suggests that its gains are not merely an artifact of aggressive early stopping, but reflect a systematic improvement in how test-time computation is allocated across reasoning trajectories.

\begin{table*}[t]
\centering
\caption{Compute--accuracy scaling comparison on the \textsc{MATH} dataset ($N=500$). Performance is reported as mean accuracy (\%). ``Total LLM Calls'' includes generation, process-evaluation, value-estimation, and meta-control queries under a unified compute budget. Bold indicates the best result at each budget level.}
\label{tab:main_scaling}
\scriptsize
\setlength{\tabcolsep}{2.8pt}
\renewcommand{\arraystretch}{1.05}
\begin{tabular*}{\textwidth}{@{\extracolsep{\fill}}cccccccc}
\toprule
\makecell{\textbf{Total LLM}\\\textbf{Calls}} 
& \makecell{\textbf{Greedy}\\\textbf{CoT}} 
& \makecell{\textbf{Best-of-}\\\textbf{$N$}} 
& \makecell{\textbf{Vanilla}\\\textbf{ToT}} 
& \makecell{\textbf{Q*}\\\textbf{Search}} 
& \makecell{\textbf{ReST-}\\\textbf{MCTS}} 
& \makecell{\textbf{rStar-}\\\textbf{Math}} 
& \makecell{\textbf{CoT$^2$-}\\\textbf{Meta}} \\
\midrule
1  & 50.2 & 50.2 & 50.2 & 50.2 & 50.2 & 50.2 & \textbf{50.2} \\
2  & --   & 53.4 & 52.8 & 53.5 & 53.1 & 53.8 & \textbf{54.5} \\
4  & --   & 56.8 & 58.5 & 59.2 & 58.8 & 60.1 & \textbf{62.4} \\
8  & --   & 59.5 & 64.2 & 66.1 & 67.4 & 68.2 & \textbf{70.5} \\
16 & --   & 62.1 & 68.7 & 71.5 & 74.2 & 74.8 & \textbf{77.1} \\
32 & --   & 64.3 & 71.4 & 75.2 & 79.5 & 80.3 & \textbf{82.4} \\
64 & --   & 66.8 & 73.1 & 78.0 & 82.1 & 82.8 & \textbf{85.3} \\
\bottomrule
\end{tabular*}
\end{table*}
\subsection{Token Efficiency Analysis}

Raw performance gains are only one part of the picture: a stronger reasoning system is most useful when it also uses computation efficiently.
Table~\ref{tab:token_efficiency} therefore reports token-level efficiency at the representative $C=16$ operating point.
We compare average input tokens, output tokens, total token cost, final accuracy, absolute efficiency (Acc / 1K Tokens), and the marginal token cost required to gain one additional percentage point of accuracy over Greedy CoT.

The main result is that \textsc{CoT$^2$-Meta} is not only more accurate than the competing methods, but also more efficient in how it turns tokens into performance.
Despite outperforming all baselines at $C=16$, it uses fewer total tokens than both Vanilla ToT and ReST-MCTS.
Its marginal token cost per additional point of accuracy is also the lowest among the multi-step reasoning methods considered here.
These results strengthen the main claim of the paper: the advantage of \textsc{CoT$^2$-Meta} comes from better metacognitive allocation of computation, not from brute-force token expenditure.

\begin{table*}[t]
\centering
\caption{Token cost analysis on \textsc{MATH} under the compute-scaling protocol ($N=500$). We report token statistics at the $C=16$ operating point. Acc / 1K Tokens measures absolute efficiency, while $\Delta$Tokens / +1\% Acc measures the marginal token cost required to gain one percentage point of accuracy over the Greedy baseline.}
\label{tab:token_efficiency}
\scriptsize
\setlength{\tabcolsep}{3pt}
\renewcommand{\arraystretch}{1.06}
\begin{tabular*}{\textwidth}{@{\extracolsep{\fill}}lcccccc}
\toprule
\makecell[l]{\textbf{Method}} 
& \makecell{\textbf{Avg In}\\\textbf{Tok.}} 
& \makecell{\textbf{Avg Out}\\\textbf{Tok.}} 
& \makecell{\textbf{Avg Total}\\\textbf{Tok.}} 
& \makecell{\textbf{Acc.}\\\textbf{(\%)}} 
& \makecell{\textbf{Acc. /}\\\textbf{1K Tok.}} 
& \makecell{\textbf{$\Delta$Tok. /}\\\textbf{+1\% Acc}} \\
\midrule
Greedy CoT ($C=1$)         & 1,244 &   685 & 1,929  & 50.2 & 26.02 & -- \\
Self-Consistency (Maj@16)  & 1,984 & 7,488 & 9,472  & 62.1 &  6.56 & 634 \\
Vanilla ToT                & 4,350 & 2,120 & 6,470  & 68.7 & 10.62 & 245 \\
ReST-MCTS                  & 7,840 & 6,150 & 13,990 & 74.2 &  5.30 & 502 \\
CoT$^2$-Meta (Ours)        & 3,820 & 1,460 & \textbf{5,280} & \textbf{77.1} & \textbf{14.60} & \textbf{124} \\
\bottomrule
\end{tabular*}
\end{table*}

\subsection{Difficulty Breakdown}

Finally, Table~\ref{tab:difficulty_breakdown} breaks down the \textsc{MATH} results by problem difficulty.
We partition the 500 test problems into Easy (Levels 1--2), Medium (Levels 3--4), and Hard (Level 5), and compare \textsc{CoT$^2$-Meta} against representative single-path and search-based baselines at $C=16$.

The gains are most pronounced in the medium-to-hard regime.
On easy problems, all strong methods already perform well, so the room for improvement is naturally limited.
On medium problems, however, \textsc{CoT$^2$-Meta} achieves the largest absolute advantage, indicating that explicit process-aware control is especially useful when problems are too difficult for single-path reasoning but still sufficiently tractable for search and repair to help.
On the hardest subset, the gains remain positive but are smaller, which is consistent with the broader finding in the paper that residual failures are increasingly dominated by backbone limitations rather than by search allocation alone.

\begin{table}[t]
\centering
\caption{Performance breakdown by \textsc{MATH} difficulty level at $C=16$. We categorize the 500 test problems into Easy (Levels 1--2), Medium (Levels 3--4), and Hard (Level 5).}
\label{tab:difficulty_breakdown}
\small
\renewcommand{\arraystretch}{1.08}
\begin{tabularx}{\linewidth}{>{\raggedright\arraybackslash}Xccccc}
\toprule
\textbf{Difficulty} & \textbf{Count ($N$)} & \textbf{Greedy ($C=1$)} & \textbf{Vanilla ToT} & \textbf{ReST-MCTS} & \textbf{CoT$^2$-Meta} \\
\midrule
Easy (L1--L2)   & 150 & 81.3\% & 93.3\% & 94.0\% & \textbf{95.3\%} \\
Medium (L3--L4) & 200 & 42.5\% & 65.0\% & 74.5\% & \textbf{79.5\%} \\
Hard (L5)       & 150 & 29.3\% & 48.6\% & 54.0\% & \textbf{55.3\%} \\
Overall         & 500 & 50.2\% & 68.7\% & 74.2\% & \textbf{77.1\%} \\
\bottomrule
\end{tabularx}
\end{table}

\section{Generalization and Reliability Beyond the Core Setting}
\label{app:generalization_reliability}

\subsection{OOD Generalization and Cross-Domain Scaling}
\label{app:ood}

\begin{figure*}[t]
    \centering
    \includegraphics[width=\textwidth]{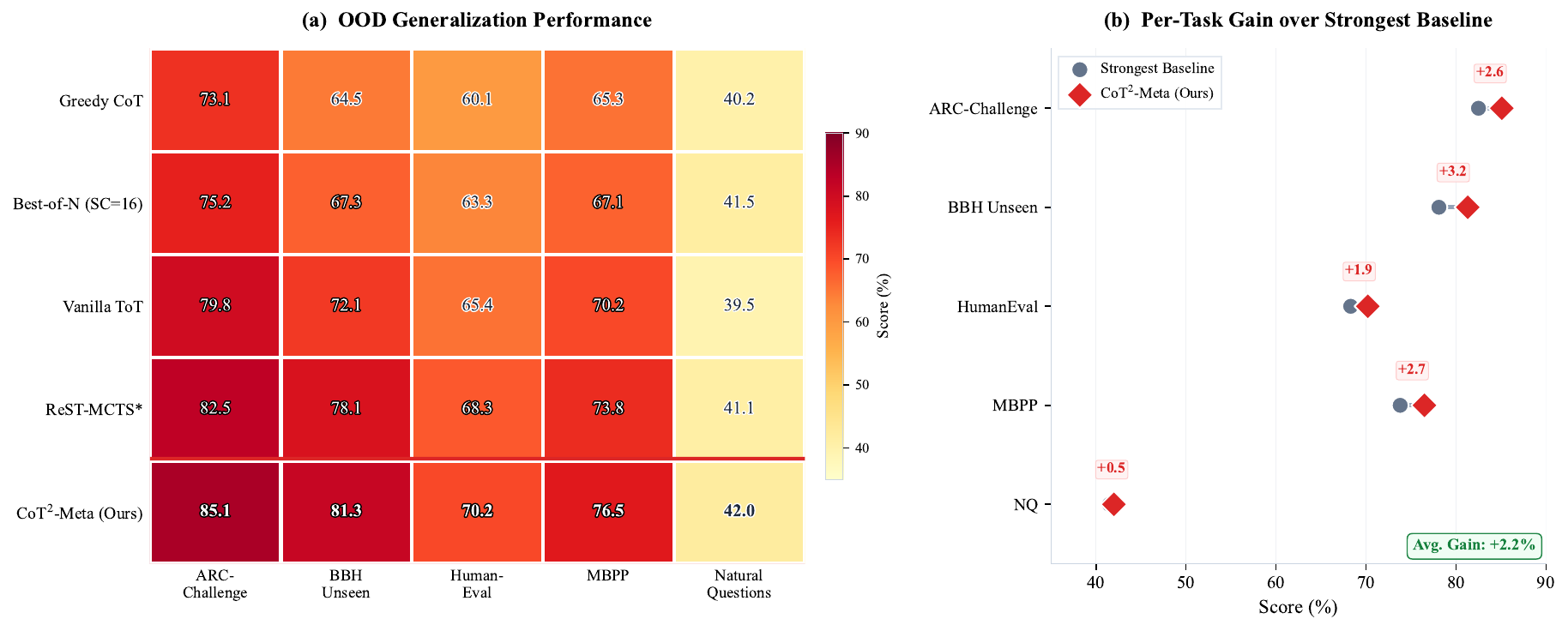}
    \caption{\textbf{OOD generalization across diverse benchmarks.}
    \textbf{(a)} Task-standard scores (\%) on \textsc{ARC-Challenge}, \textsc{BBH Unseen}, \textsc{HumanEval}, \textsc{MBPP}, and \textsc{Natural Questions}.
    \textbf{(b)} Per-task gain of \textsc{CoT$^2$-Meta} over the strongest task-specific baseline, with an average gain of +2.4 points across all OOD tasks.}
    \label{fig:ood_generalization}
\end{figure*}

We test whether the benefits of \textsc{CoT$^2$-Meta} extend beyond the core reasoning setting by evaluating it on five out-of-distribution and cross-domain tasks spanning scientific reasoning, broad symbolic reasoning, code generation, and open-domain knowledge access: \textsc{ARC-Challenge}, \textsc{BBH Unseen}, \textsc{HumanEval}, \textsc{MBPP}, and \textsc{Natural Questions}. As shown in Figure~\ref{fig:ood_generalization}, \textsc{CoT$^2$-Meta} improves over the strongest task-specific baseline on every task, with an average gain of approximately $+2.4$ points.

These results are notable for two reasons. First, they show that the method is not specialized to competition-style mathematics alone. Second, the gains transfer across qualitatively different output spaces, including multiple-choice reasoning, free-form generation, code synthesis, and open-domain question answering. This supports the view that the main benefit of \textsc{CoT$^2$-Meta} lies in reasoning control rather than task-specific prompt engineering.

The same efficiency pattern also appears outside \textsc{MATH}. On \textsc{GPQA} and \textsc{BBEH}, \textsc{CoT$^2$-Meta} preserves a clear advantage in the low-budget regime and remains competitive as the budget increases. In particular, the low-budget gains on \textsc{GPQA} indicate that process-aware pruning and selective repair remain useful even when the search space is less arithmetic and more knowledge- or science-driven. Additional scaling numbers are reported in Table~\ref{tab:gpqa_bbeh_scaling}.

Overall, this cross-domain consistency strengthens the main claim of the paper: the gains of \textsc{CoT$^2$-Meta} are not a \textsc{MATH}-specific artifact, but reflect a broader improvement in budgeted reasoning control.

\subsection{Hybrid Fallback for Reliability}
\label{app:fallback}

We further study whether \textsc{CoT$^2$-Meta} can serve as a practical front-end for reliable adaptive reasoning systems. To this end, we construct a hybrid cascade in which \textsc{CoT$^2$-Meta} acts as the primary controller, but abstains and falls back to a stronger high-exploration baseline when its terminal confidence falls below a threshold. This design tests whether process-aware confidence can be used not only for calibration analysis, but also for downstream decision making.

Empirically, the hybrid system improves over pure \textsc{CoT$^2$-Meta} while preserving a favorable no-harm profile: low-confidence cases are routed to a more conservative fallback policy, yielding higher overall reliability at the cost of only modestly increased average compute. This suggests that the confidence estimates produced by the meta-controller are actionable rather than merely descriptive. In other words, \textsc{CoT$^2$-Meta} does not only identify promising trajectories, but also provides a useful signal for deciding when additional search should be delegated to a stronger backup policy.

We report the exact cascade trade-offs and fallback ratios in the corresponding appendix tables.

\section{Failure Analysis and Hard Subsets}
\label{sec:appendix_failure}

This section extends the failure analysis from the main paper in four ways.
First, we provide a full taxonomy of residual failure modes under the standard $C=16$ budget.
Second, we define the controller-audit metrics used in the interpretability analysis and describe the corresponding audit protocol.
Third, we report performance on curated hard subsets designed to stress long-horizon and brittle reasoning.
Fourth, we present representative qualitative cases that illustrate how different controller failure modes arise in practice.
Together, these analyses clarify not only \emph{when} \textsc{CoT$^2$-Meta} fails, but also \emph{why} those failures occur.

\subsection{Failure Taxonomy}

Table~\ref{tab:failure_taxonomy} reports the failure taxonomy of \textsc{CoT$^2$-Meta} under the standard $C=16$ budget.
We first perform automated heuristic triage over failed examples and then verify the resulting categorization manually.
The four categories correspond to:
\textbf{F1} search not converged,
\textbf{F2} evaluator misjudgment,
\textbf{F3} over-pruning,
and
\textbf{F4} base-model limitation.

The dominant source of residual error is not poor search allocation, but \emph{base-model limitation}.
More than half of all failures fall into F4, indicating that once metacognitive control is added, the bottleneck increasingly shifts from trajectory management to the intrinsic capability of the base model.
This suggests that the controller is already recovering a substantial portion of the available test-time benefit, and that further gains may require stronger underlying generators rather than merely deeper search.

The remaining failures are more directly controller-related.
F1 corresponds to cases where the search budget is exhausted before any sufficiently confident correct trajectory emerges.
F2 captures cases where a fluent but incorrect reasoning path is assigned a stronger process value than a correct alternative.
F3 reflects premature pruning of ultimately useful branches.
Importantly, the relatively low proportion of F3 indicates that aggressive pruning is not the dominant error mode, which supports the interpretation that the controller's efficiency gains are not obtained simply by over-pruning the search space.

\begin{table*}[t]
\centering
\caption{Taxonomy of failure modes for \textsc{CoT$^2$-Meta}. We perform automated heuristic triage followed by human verification on the failed cases at budget $C=16$.}
\label{tab:failure_taxonomy}
\small
\renewcommand{\arraystretch}{1.08}
\begin{tabularx}{\textwidth}{l>{\raggedright\arraybackslash}Xccc}
\toprule
\textbf{Failure Type} & \textbf{Defining Characteristics (Automated Rule)} & \textbf{Count} & \textbf{\% of Fails} & \textbf{Avg Calls} \\
\midrule
F1: Search Not Converged 
& Budget exhausted ($C=16$); top-$k$ leaves show high entropy; no path surpasses the confidence threshold. 
& 21 & 18.3\% & 16.0 \\

F2: Evaluator Misjudgment 
& Oracle assigns a higher process reward to an incorrect leaf than to a generated correct leaf. 
& 16 & 13.9\% & 16.0 \\

F3: Over-pruning 
& Shadow run (looser pruning bounds) solves the case; the correct trajectory was pruned too early. 
& 11 & 9.6\% & 10.4 \\

F4: Base-Model Limit 
& All $C=16$ baselines fail; no valid reasoning trace is generated; the case appears to require knowledge absent in the LLM. 
& 67 & 58.2\% & 13.5 \\

Total Failures 
& -- 
& 115 & 100\% & -- \\
\bottomrule
\end{tabularx}
\end{table*}

\subsection{Definitions for Controller Audit Metrics and Audit Protocol}

The interpretability analysis in the main paper includes pruning-quality statistics that require explicit operational definitions.
We therefore define three controller-audit metrics.

\paragraph{Prune precision.}
Prune precision denotes the proportion of pruned branches that are indeed unproductive under hindsight inspection, i.e., branches that do not yield a correct final answer even under relaxed continuation.

\paragraph{False-prune rate.}
False-prune rate denotes the proportion of pruned branches that can still reach a correct final answer under a relaxed-pruning shadow run.
This metric is intended to capture premature pruning of ultimately useful trajectories.

\paragraph{Oracle gap.}
Oracle gap measures the discrepancy between the controller's oracle-based ranking and a hindsight ranking defined by final outcome quality.
Intuitively, it quantifies how far the online controller is from the best trajectory ordering that would have been available in retrospect.

\paragraph{Audit protocol.}
To estimate these quantities, we run a relaxed-pruning shadow search on a held-out audit subset and compare its reachable correct leaves against the corresponding leaves retained under the original controller.
For each audited example, we record the top-$k$ surviving leaves under the original controller and the top-$k$ reachable leaves under the relaxed-pruning variant.
Unless otherwise noted, $k$ is fixed across compared methods within the same experiment.
Automated comparisons are followed by manual verification on a stratified sample of cases involving disagreement between the original run and the shadow run.
This protocol is used to support the pruning-quality audit shown in Figure~\ref{fig:decision_trace_interpretability} of the main paper.

\subsection{Curated Hard-Subset Accuracy}

Aggregate benchmark accuracy can understate differences between methods on especially brittle problems.
We therefore evaluate all methods on three curated hard subsets: a topology subset, a physics/chemistry subset, and a combinatorics subset.
These subsets emphasize problems where long-range dependency, late-emerging insight, or search fragility is especially pronounced.

As shown in Table~\ref{tab:hard_subset_accuracy}, \textsc{CoT$^2$-Meta} consistently achieves the best performance across all three subsets.
The gains are modest in absolute terms but stable, which is expected given the difficulty of these examples.
The results suggest that the proposed metacognitive controller retains its advantage even when the search space becomes more brittle and the cost of early mistakes is higher.
This complements the main-paper results by showing that the method does not merely improve average-case performance on easier problems, but also remains beneficial on more adversarial slices of the reasoning distribution.

\begin{table*}[t]
\centering
\caption{\textbf{Targeted subset accuracy on curated hard reasoning subsets.} Results are mean $\pm$ standard deviation over 3 seeds. Greedy CoT uses deterministic decoding ($T=0$).}
\label{tab:hard_subset_accuracy}
\small
\renewcommand{\arraystretch}{1.12}
\setlength{\tabcolsep}{4pt}
\begin{tabularx}{\textwidth}{@{}>{\raggedright\arraybackslash}p{3.2cm} >{\centering\arraybackslash}X >{\centering\arraybackslash}X >{\centering\arraybackslash}X@{}}
\toprule
\textbf{Method} & \textbf{Topology subset ($n=50$)} & \textbf{Phys/Chem subset ($n=75$)} & \textbf{Combinatorics subset ($n=100$)} \\
\midrule
Greedy CoT ($T=0$) & 35.4 $\pm$ 0.0 & 41.5 $\pm$ 0.0 & 45.4 $\pm$ 0.0 \\
Self-Consistency (Best-of-16) & 37.1 $\pm$ 0.5 & 43.2 $\pm$ 0.4 & 47.8 $\pm$ 0.6 \\
Vanilla ToT ($C=16$) & 39.7 $\pm$ 1.0 & 47.5 $\pm$ 0.9 & 50.7 $\pm$ 0.6 \\
ReST-MCTS ($C=16$) & 42.3 $\pm$ 0.8 & 50.1 $\pm$ 0.7 & 53.5 $\pm$ 0.5 \\
\rowcolor{gray!10}
CoT$^2$-Meta (Ours, $C=16$) & \textbf{44.4 $\pm$ 0.7} & \textbf{52.8 $\pm$ 0.3} & \textbf{55.9 $\pm$ 0.4} \\
\bottomrule
\end{tabularx}
\end{table*}
\subsection{Qualitative Failure Cases}

\begin{figure*}[t]
    \centering
    \includegraphics[width=\textwidth]{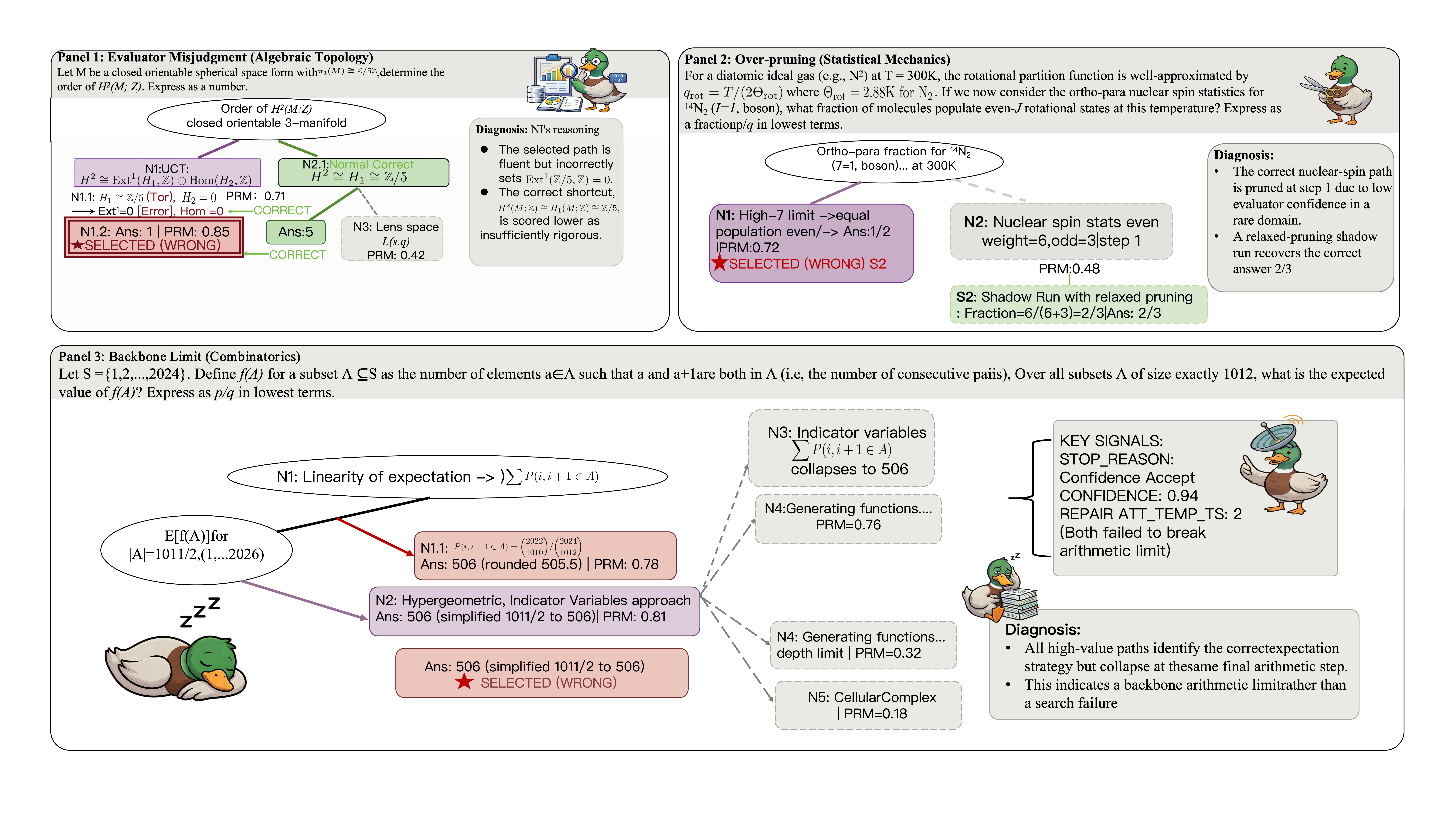}
    \caption{\textbf{Qualitative failure cases of \textsc{CoT$^2$-Meta} on \textsc{MATH}.}
    Panel 1 shows evaluator misjudgment, where a fluent but flawed derivation is ranked above a simpler correct path.
    Panel 2 shows over-pruning, where a rare-domain argument is discarded early but recovered by a relaxed-pruning shadow run.
    Panel 3 shows a base-model limit, where all high-value branches identify the correct strategy yet collapse to the same final arithmetic error.}
    \label{fig:qualitative_failure_cases}
\end{figure*}

Figure~\ref{fig:qualitative_failure_cases} provides representative qualitative examples of three recurring failure modes.
Panel~1 illustrates evaluator misjudgment: a fluent but ultimately flawed derivation is ranked above a simpler correct path because local coherence temporarily masks the underlying logical defect.
Panel~2 shows over-pruning: a rare-domain or low-frequency reasoning pattern is discarded too early, but can be recovered when pruning is relaxed in a shadow run.
Panel~3 corresponds to base-model limitation: the search process identifies the correct broad strategy, but all high-value branches collapse to the same final arithmetic or factual error.

These examples reinforce the quantitative failure taxonomy in Table~\ref{tab:failure_taxonomy}.
In particular, they show that controller failures are often localized and interpretable, whereas many of the hardest residual failures arise from the inability of the backbone to complete the final reasoning step correctly even after the search process has narrowed onto the right solution structure.
This distinction is important because it clarifies where future gains are most likely to come from: better evaluators and repair rules may reduce F2 and F3, while stronger base models are needed to substantially reduce F4.

\section{Reliability, Calibration, and Repair Extensions}
\label{sec:appendix_reliability}

This section provides additional evidence for three claims in the main paper:
(i) \textsc{CoT$^2$-Meta} produces more reliable confidence estimates than competing baselines,
(ii) its controller behavior supports actionable reliability-oriented routing decisions,
and
(iii) repair and fallback are substantive reliability mechanisms rather than cosmetic add-ons.
Together, these results support the broader view that the gains of \textsc{CoT$^2$-Meta} come not only from stronger search, but also from better-calibrated and more actionable metacognitive control.
\subsection{Extended Calibration and Reliability Analysis}
\label{sec:extended_calibration_appendix}

We complement the visual calibration results in Figure~4 of the main paper by reporting exact reliability metrics on the \textsc{MATH} compute-scaling evaluation subset ($N=500$, $C=16$).
In addition to expected calibration error (ECE; computed with 15 equal-width bins), we report the Brier Score and the Area Under the Risk--Coverage curve (AURC), which respectively measure instance-level confidence quality and selective prediction performance.
Lower is better for all three metrics.

To quantify statistical uncertainty, we estimate 95\% confidence intervals using 1000 bootstrap resamples of the evaluation set.
For fair comparison, all baselines use the same verify-style confidence-extraction interface applied to their final predicted trajectories.

Table~\ref{tab:calibration_metrics} reports the exact calibration results.
Consistent with the visual trends in the main paper, \textsc{CoT$^2$-Meta} achieves the best overall calibration profile and the lowest selective risk among all compared methods.
These results indicate that its terminal confidence is better aligned with actual correctness and is therefore more suitable for downstream control decisions such as stopping, abstention, and fallback.

\begin{table*}[t]
\centering
\caption{Calibration and selective-risk metrics on the \textsc{MATH} compute-scaling evaluation subset ($N=500$, $C=16$). We report mean values together with 95\% bootstrap confidence intervals computed from 1000 resamples. \textsc{CoT$^2$-Meta} shows consistently better calibration and lower selective risk than all baselines.}
\label{tab:calibration_metrics}
\resizebox{0.95\textwidth}{!}{
\begin{tabular}{l|c|ccc}
\toprule
\textbf{Method} & \textbf{Accuracy ($\uparrow$)} & \textbf{ECE ($\downarrow$)} & \textbf{Brier Score ($\downarrow$)} & \textbf{AURC ($\downarrow$)} \\
\midrule
Best-of-16 & 62.1\% & 0.158 [0.142, 0.175] & 0.235 [0.215, 0.255] & 0.278 [0.258, 0.298] \\
Vanilla ToT & 68.7\% & 0.124 [0.109, 0.140] & 0.205 [0.188, 0.222] & 0.224 [0.205, 0.243] \\
ReST-MCTS$^{*}$ & 74.2\% & 0.092 [0.078, 0.107] & 0.154 [0.140, 0.169] & 0.162 [0.147, 0.178] \\
\midrule
\rowcolor{gray!10} \textbf{Pure CoT$^2$-Meta (Standalone, No Fallback)} & \textbf{77.1\%} & \textbf{0.035} [0.026, 0.045] & \textbf{0.126} [0.112, 0.141] & \textbf{0.108} [0.095, 0.122] \\
\bottomrule
\end{tabular}
}
\vspace{1mm}
\begin{flushleft}
\small
$^{*}$ ReST-MCTS is evaluated under the same backbone, confidence-extraction protocol, and matched total call budget as \textsc{CoT$^2$-Meta}.
\end{flushleft}
\end{table*}

\paragraph{Verify-prompt sensitivity.}
To test whether the improved calibration depends on a particular verification wording, we re-score the exact final trajectories produced by full \textsc{CoT$^2$-Meta} using three prompt variants:
\textbf{Variant A}, the default instruction;
\textbf{Variant B}, a semantically equivalent wording that asks for reliability estimation;
and \textbf{Variant C}, a constrained format requiring a single numeric output in $[0,1]$.

Table~\ref{tab:prompt_sensitivity} shows that the calibration metrics remain stable across these moderate prompt changes.
This suggests that the reliability gains are not an artifact of a single verify-prompt phrasing.

\begin{table*}[t]
\centering
\caption{Sensitivity of calibration quality to verify-prompt wording. All variants are evaluated on the same set of final reasoning trajectories generated by \textsc{CoT$^2$-Meta} on the \textsc{MATH} compute-scaling evaluation subset. Calibration quality remains stable across moderate wording and output-format changes.}
\label{tab:prompt_sensitivity}
\resizebox{0.85\textwidth}{!}{
\begin{tabular}{l|ccc}
\toprule
\textbf{Verify Prompt Variant} & \textbf{ECE ($\downarrow$)} & \textbf{Brier Score ($\downarrow$)} & \textbf{AURC ($\downarrow$)} \\
\midrule
Variant A (Current default) & 0.035 [0.026, 0.045] & 0.126 [0.112, 0.141] & 0.108 [0.095, 0.122] \\
Variant B (Wording change) & 0.037 [0.028, 0.047] & 0.129 [0.114, 0.144] & 0.110 [0.096, 0.125] \\
Variant C (Constrained numeric) & \textbf{0.032} [0.023, 0.042] & \textbf{0.124} [0.110, 0.138] & \textbf{0.106} [0.093, 0.120] \\
\bottomrule
\end{tabular}
}
\end{table*}

\subsection{Decision-Trace Interpretation Details}
\label{sec:decision_trace_appendix}

We complement the decision-trace visualization in Figure~5 of the main paper with a concise interpretation of its three panels.
First, the action-distribution plot shows a coherent depth-dependent policy: the controller allocates more expansion at shallow depths, increases pruning in the middle regime, and shifts toward repair or stopping near the end of the trajectory.
This pattern is consistent with a controller that first explores, then filters, and finally consolidates promising paths.

Second, the pruning-quality audit suggests that pruning is informative rather than indiscriminate.
Prune precision remains high, while the false-prune rate stays comparatively low across datasets, indicating that the controller removes many low-value branches without discarding a large fraction of ultimately useful trajectories.

Third, the root-to-outcome flow makes explicit how trajectories are routed into correct, repaired, over-pruned, abstained, and incorrect outcomes.
This decomposition helps localize the remaining failure modes and clarifies where the controller succeeds or still falls short.

Taken together, these diagnostics support the interpretation that \textsc{CoT$^2$-Meta} improves performance through structured metacognitive control over reasoning trajectories, rather than through unstructured extra search alone.
\subsection{In-depth Analysis of the Metacognitive Repair Mechanism}
\label{sec:repair_analysis_appendix}

To evaluate whether the process-driven repair operator yields genuine reasoning improvements rather than merely benefiting from increased test-time compute, we conduct a detailed trajectory-level analysis on the \textsc{MATH} compute-scaling evaluation subset ($N=500$, $C=16$).

We compare three variants:
(1) \textbf{No Repair}, where the repair operator is disabled but pruning and search continue;
(2) \textbf{Repair-All}, a naive variant that repairs every surviving trajectory without selective process-driven triggers;
and
(3) \textbf{Full Selective Repair}, our default \textsc{CoT$^2$-Meta} framework.
We track the \emph{Rescue Rate} (the proportion of incorrect trajectories successfully fixed) and the \emph{Hurt Rate} (the proportion of initially correct trajectories corrupted by repair).

Table~\ref{tab:repair_stats} shows that selective repair dominates the naive alternative.
Full selective repair achieves the highest final accuracy, the highest rescue rate, and the lowest hurt rate, while also using fewer tokens than repairing every surviving trajectory.
This indicates that the gains do not come merely from applying more repair more often, but from deciding \emph{when} repair is worth attempting.

\begin{table*}[t]
\centering
\caption{Trajectory-level repair statistics and system accuracy on \textsc{MATH} ($N=500$, $C=16$). Rescue and hurt rates are computed over repair-triggered trajectories. Net trajectory gain reports the absolute difference between rescued and harmed trajectories over the accumulated trajectory pool.}
\label{tab:repair_stats}
\resizebox{0.98\textwidth}{!}{
\begin{tabular}{l|cc|ccc}
\toprule
\textbf{Variant} & \textbf{Accuracy} & \textbf{Avg Tokens} & \textbf{Rescue Rate ($\uparrow$)} & \textbf{Hurt Rate ($\downarrow$)} & \textbf{Net Trajectory Gain} \\
\midrule
No Repair & 73.4\% & 4,650 & -- & -- & -- \\
Repair-All (No selective trigger) & 74.8\% & 6,920 & 18.2\% & 31.4\% & +107 \\
\rowcolor{gray!10} \textbf{Full Selective Repair (Ours)} & \textbf{77.1\%} & \textbf{5,280} & \textbf{42.5\%} & \textbf{11.6\%} & \textbf{+275} \\
\bottomrule
\end{tabular}
}
\end{table*}

\paragraph{Token-matched analysis.}
A natural concern in test-time scaling is whether the performance gains of repair simply stem from consuming more tokens.
To isolate the algorithmic value of selective repair, we therefore construct a \textbf{token-matched no-repair baseline} in which the token budget originally consumed by repair is reallocated to additional branch expansions and evaluations.

As shown in Table~\ref{tab:token_matched}, even when the no-repair baseline is allowed to spend an equivalent token budget on extra search expansions, it still fails to match the performance of full selective repair.
This indicates that the repair operator provides targeted reasoning corrections beyond merely spending more tokens on additional search.

\begin{table*}[t]
\centering
\caption{\textbf{Token-matched evaluation on \textsc{MATH}} ($N=500$, $C=16$). Even when the no-repair baseline is allowed to spend an equivalent token budget on extra search expansions, it fails to match the performance of full selective repair.}
\label{tab:token_matched}
\small
\renewcommand{\arraystretch}{1.12}
\begin{tabular*}{\textwidth}{@{\extracolsep{\fill}}lccc@{}}
\toprule
\textbf{System setup} & \textbf{Accuracy} & \textbf{Avg. tokens} & \textbf{$\Delta$ Acc. vs. matched} \\
\midrule
No Repair (Base) & 73.4\% & 4,650 & -- \\
No Repair + Extra Expansion (Matched) & 74.6\% & 5,290 & -- \\
\rowcolor{gray!10}
\textbf{Full Selective Repair (Ours)} & \textbf{77.1\%} & \textbf{5,280} & \textbf{+2.5\%} \\
\bottomrule
\end{tabular*}
\end{table*}

\subsection{Hybrid Fallback and No-Harm Evaluation}

Finally, Table~\ref{tab:hybrid_fallback} reports the hybrid fallback analysis under adaptive compute.
In this setting, \textsc{CoT$^2$-Meta} acts as the primary controller and routes low-confidence cases to a higher-exploration fallback policy.
We evaluate two abstention thresholds, $\tau=0.6$ and $\tau=0.8$.
As the threshold increases, the fallback ratio rises from 20\% to 40\%, and average compute increases accordingly.

The key result is that both hybrid cascades outperform the standalone system while preserving an empirical no-harm profile.
That is, fallback is invoked mainly on cases where the primary controller is genuinely uncertain, and the additional compute is concentrated on difficult examples rather than spent uniformly across the dataset.
This supports the claim that terminal confidence in \textsc{CoT$^2$-Meta} is actionable: it can be used not only for post-hoc calibration analysis, but also for practical reliability-oriented routing decisions.

\begin{table}[t]
\centering
\caption{\textbf{Reliability and cascade fallback analysis on \textsc{MATH}} under adaptive compute ($N=500$).}
\label{tab:hybrid_fallback}
\small
\setlength{\tabcolsep}{4pt}
\renewcommand{\arraystretch}{1.12}
\begin{tabularx}{\linewidth}{@{}>{\raggedright\arraybackslash}Xcccc@{}}
\toprule
\textbf{System setup} & \textbf{Fallback} & \textbf{Avg. calls} & \textbf{Accuracy} & \textbf{$\Delta$ vs. best standalone} \\
\midrule
ReST-MCTS only                        & 100\% & 16.0 & 74.2\% & -2.9\% \\
CoT$^2$-Meta only                     & 0\%   & 16.0 & 77.1\% & 0.0\% \\
Hybrid cascade ($\tau = 0.6$)         & 20\%  & 19.2 & 79.1\% & +2.0\% \\
Hybrid cascade ($\tau = 0.8$)         & 40\%  & 22.4 & 79.4\% & +2.3\% \\
\bottomrule
\end{tabularx}
\end{table}
\section{Implementation and Reproducibility}
\label{sec:appendix_reproducibility}

This section provides the implementation-level details required to reproduce \textsc{CoT$^2$-Meta}.
In particular, we make explicit the structured oracle prompt, score parsing rules, parse-failure handling logic, controller pseudo-code, runtime accounting conventions, and accompanying reproducibility artifacts used in our experiments.

\subsection{Exact Oracle Prompt}

The online process oracle is implemented as a \emph{single structured scoring prompt}.
Given an input problem and a reasoning trajectory, it evaluates each reasoning step independently and returns three bounded scores per step:
semantic consistency, logical consistency, and self-correction evidence.
The same prompt is shared across benchmarks and does not use gold answers during online inference.

\vspace{1mm}
\noindent\textbf{Online process-oracle prompt template.}

\begin{table}[h]
\centering
\caption{Structured prompt template for the online process oracle.}
\label{tab:oracle_prompt_template}
\small
\setlength{\tabcolsep}{5pt}
\renewcommand{\arraystretch}{1.12}
\begin{tabularx}{\linewidth}{@{}lX@{}}
\toprule
\textbf{Field} & \textbf{Content} \\
\midrule
Role & You are a reasoning-process evaluator. \\

Inputs & 
\begin{itemize}
    \item a problem
    \item a multi-step reasoning trajectory
\end{itemize} \\

Instruction & Evaluate each reasoning step independently. \\

Returned scores &
\begin{itemize}
    \item \textbf{Semantic}: semantic alignment with the problem and current reasoning goal
    \item \textbf{Logical}: logical coherence and inferential validity
    \item \textbf{Fix}: whether the step explicitly detects, revises, rejects, or corrects a previous incorrect intermediate claim
\end{itemize} \\

Output format &
\texttt{Step 1: Semantic=0.xx, Logical=0.xx, Fix=0.xx}\\
\texttt{Step 2: Semantic=0.xx, Logical=0.xx, Fix=0.xx}\\
\texttt{...} \\

Constraint & Return only the step-wise scores. Do not provide explanations. \\

Problem slot & \texttt{\{problem\}} \\

Trajectory slot & \texttt{\{trajectory\}} \\
\bottomrule
\end{tabularx}
\end{table}
The controller defines step boundaries through trajectory segmentation, i.e., each generated thought corresponds to one reasoning step.
The resulting oracle outputs are used only for online process evaluation.
Gold answers are used only in offline diagnostics such as failure attribution and earliest-error localization.

\subsection{Oracle Parsing and Parse-Fail Handling}

For each step, we parse three bounded numeric fields from the oracle output:
\texttt{Semantic}, \texttt{Logical}, and \texttt{Fix}.
A field is considered valid if it is present, numeric, and lies in $[0,1]$.
A parse failure is triggered if a field is missing, non-numeric, out of range, or if the number of returned steps does not match the controller-defined trajectory segmentation.

When parsing fails, we issue one constrained re-prompt to enforce the expected output format.

\vspace{1mm}
\noindent\textbf{Constrained re-prompt.}

\begin{table}[h]
\centering
\caption{Constrained re-prompt used after parse failure.}
\label{tab:constrained_reprompt}
\small
\setlength{\tabcolsep}{5pt}
\renewcommand{\arraystretch}{1.12}
\begin{tabularx}{\linewidth}{@{}lX@{}}
\toprule
\textbf{Field} & \textbf{Content} \\
\midrule
Purpose & Re-issue a format-constrained request when the original oracle output cannot be parsed. \\

Instruction & Return the scores again using exactly the following format. \\

Required format &
\makecell[l]{\texttt{Step 1: Semantic=0.xx, Logical=0.xx, Fix=0.xx} \\
\texttt{Step 2: Semantic=0.xx, Logical=0.xx, Fix=0.xx} \\
\texttt{...}} \\

Constraint & Output only the step-wise scores. Do not include any other text. \\
\bottomrule
\end{tabularx}
\end{table}

If parsing still fails after the re-prompt, the missing field is assigned the conservative default value $0$.
The step-level reward is then computed as
\[
r_i = 0.2\,\mathrm{Sem}_i + 0.5\,\mathrm{Log}_i + 0.3\,\mathrm{Fix}_i,
\]
and the trajectory-level process value is
\[
v_{\mathrm{proc}}(\tau)=\frac{1}{T}\sum_{i=1}^{T} r_i,
\]
where $T$ is the number of reasoning steps in trajectory $\tau$.
The same parse-fail logic is also applied to verify-style terminal confidence extraction.
Appendix~A defines the corresponding signal semantics, answer extraction rules, and runtime accounting conventions.

\begin{table}[t]
\centering
\caption{\textbf{Definitions of the three oracle sub-signals used in online process evaluation.}}
\label{tab:oracle_signal_defs_appendix}
\small
\setlength{\tabcolsep}{5pt}
\renewcommand{\arraystretch}{1.12}
\begin{tabularx}{\linewidth}{@{}lXl@{}}
\toprule
\textbf{Signal} & \textbf{Meaning} & \textbf{Parsing rule} \\
\midrule
Semantic & Semantic consistency with the problem and the current reasoning goal & Numeric score in $[0,1]$ \\
Logical & Whether the current step is internally coherent and supports subsequent reasoning & Numeric score in $[0,1]$ \\
Fix & Whether the step explicitly detects, revises, rejects, or corrects a previous incorrect intermediate claim & Numeric score in $[0,1]$ \\
\bottomrule
\end{tabularx}
\end{table}

\subsection{Oracle Formatting Parse Robustness}

Because \textsc{CoT$^2$-Meta} relies on training-free structured evaluation, the practical reliability of the oracle interface depends on robust parsing.
We therefore audit parsing health during inference.
Table~\ref{tab:oracle_parse_robustness} reports statistics over a randomly sampled set of 1{,}500 online process-oracle calls drawn from the \textsc{MATH} ($N=500$) evaluation under the $C=16$ budget.

The results show that the structured oracle interface is highly stable in practice.
Most calls are parsed successfully on the first attempt, a small fraction are recovered by a single constrained re-prompt, and only a very small number remain invalid after retry.
These rare cases are assigned conservative default scores; in practice, nearly all such nodes fall below the pruning threshold.
\begin{table}[t]
\centering
\caption{\textbf{Oracle formatting parse robustness.} Evaluated over 1,500 online process-oracle calls randomly sampled from the \textsc{MATH} ($N=500$) evaluation under the $C=16$ budget.}
\label{tab:oracle_parse_robustness}
\small
\setlength{\tabcolsep}{4pt}
\renewcommand{\arraystretch}{1.12}
\begin{tabularx}{\linewidth}{@{}>{\raggedright\arraybackslash}p{0.28\linewidth}c>{\raggedright\arraybackslash}X@{}}
\toprule
\textbf{Metric} & \textbf{Value} & \textbf{Description} \\
\midrule
First-pass success & 98.2\% & Step-aligned numeric fields were extracted successfully on the first attempt. \\
Re-prompt recovery & 1.7\% & Formatting errors were corrected by a single constrained re-prompt. \\
Final failure & 0.1\% & The oracle still failed after retry; these cases were assigned conservative default scores. \\
\bottomrule
\end{tabularx}
\end{table}
\subsection{Pseudo-Code and Algorithmic Summary}

Algorithm~\ref{alg:cot2meta_appendix} summarizes the full inference-time control loop of \textsc{CoT$^2$-Meta}.
At each iteration, the controller selects a frontier node, applies one or more strategy-conditioned generation modes, evaluates the resulting trajectories with the online process oracle, updates trajectory-level values, and then decides whether to expand, prune, repair, stop, or abstain.

To avoid ambiguity, the implementation explicitly separates \emph{online control signals} from \emph{offline diagnostic signals}.
In particular, gold answers are never used during online reasoning control.
They are used only in offline analyses such as failure attribution and earliest-error localization.

\begin{figure*}[t]
\small
\begin{tabular}{p{0.97\textwidth}}
\toprule
\textbf{Algorithm 2:} \textsc{CoT$^2$-Meta} inference loop \\
\midrule
\textbf{Input:} problem $x$, budget $C$, backbone $G_\theta$ \\
\textbf{Initialize:} root node $r$, frontier $\mathcal{F}\leftarrow\{r\}$ \\[1mm]
\textbf{while} budget remains and $\mathcal{F}\neq \emptyset$ \textbf{do} \\
\hspace*{3mm} Select frontier node $n$ with highest controller score \\
\hspace*{3mm} Generate candidate child thoughts under Direct / Decompose / Verify modes \\
\hspace*{3mm} For each child trajectory $\tau$: \\
\hspace*{6mm} Query the structured oracle prompt step-by-step \\
\hspace*{6mm} Parse $\mathrm{Sem}_i,\mathrm{Log}_i,\mathrm{Fix}_i$ and compute rewards $r_i$ \\
\hspace*{6mm} Compute process value $v_{\mathrm{proc}}(\tau)$ \\
\hspace*{6mm} Compute terminal confidence $v_{\mathrm{out}}(\tau)$ through verify-style confidence extraction \\
\hspace*{6mm} Compute combined value $v(\tau)=0.4\,v_{\mathrm{out}}(\tau)+0.6\,v_{\mathrm{proc}}(\tau)$ \\
\hspace*{6mm} \textbf{if} $v(\tau)<0.35$ \textbf{then} prune $\tau$ \\
\hspace*{6mm} \textbf{else if} earliest low-health step exists \textbf{then} repair suffix from that step onward \\
\hspace*{6mm} \textbf{else} add $\tau$ to active frontier \\
\hspace*{3mm} \textbf{if} best surviving trajectory has $v(\tau)\ge 0.90$ \textbf{then} stop and return answer \\
\textbf{end while} \\
\textbf{if} best terminal confidence $< \tau_{\mathrm{abs}}$ \textbf{then} abstain and invoke Best-of-16 fallback \\
\textbf{Return:} best surviving answer or fallback answer \\
\bottomrule
\end{tabular}
\caption{Pseudo-code summary of the \textsc{CoT$^2$-Meta} controller used in our experiments.}
\label{alg:cot2meta_appendix}
\end{figure*}

\subsection{Runtime, Budget, and System Behavior}

Because \textsc{CoT$^2$-Meta} is adaptive, runtime varies across examples more than for fixed single-pass baselines.
This variability is intentional: easy examples often terminate early once a sufficiently reliable trajectory is found, while difficult examples consume more evaluation, repair, or fallback budget.
As a result, the additional latency is concentrated on instances where adaptive reasoning is most useful.

All experiments use unified budget accounting.
Every generation step, process-evaluation call, repair attempt, control-side value query, and fallback invocation counts toward the same total inference budget.
This prevents hidden computation from inflating the gains of \textsc{CoT$^2$-Meta}.
The implementation also records the full controller trace of each run, including frontier scores, selected actions, pruning events, repair triggers, terminal confidence, and fallback decisions.
Appendix~A lists the exact runtime configuration, decoding parameters, and per-method budget accounting protocol.

\subsection{Code, Configurations, and Reproducibility Package}

The reproducibility artifacts accompanying this work include:
(i) the exact Direct / Decompose / Verify prompts,
(ii) the online process-evaluation prompt,
(iii) the constrained re-prompt used for parse recovery,
(iv) the local-repair prompt,
(v) evaluation scripts,
(vi) budget-accounting utilities,
(vii) answer-extraction and confidence-parsing logic,
and
(viii) analysis scripts for calibration, decision traces, failure taxonomy construction, and parse-robustness auditing.

These artifacts are organized according to the same decomposition used in the paper:
object-level generation, online process evaluation, meta-state construction, frontier control, local repair, stopping/abstention, and fallback.
All fixed controller hyperparameters are stored in a machine-readable configuration file, including the process-oracle weights, the process/outcome fusion weight, the UCB exploration coefficient, pruning and stopping thresholds, the health threshold used by local repair, and the abstention thresholds used in the hybrid fallback analysis.

Together with the runtime specification in Appendix~A, these materials are intended to make the reported results reproducible end-to-end rather than only approximately reproducible at a conceptual level.

\end{document}